\documentclass[11pt]{article}

\usepackage{acl}

\usepackage{times}
\usepackage{latexsym}
  \usepackage{booktabs}
  \usepackage{multirow}
  \usepackage{colortbl}
  \usepackage{xcolor}
  \usepackage[most]{tcolorbox}
 \usepackage[normalem]{ulem}
 \usepackage{titletoc}
\usepackage[T1]{fontenc}
\usepackage{enumitem}
\usepackage{dsfont}
\usepackage{paralist}

\usepackage[utf8]{inputenc}

\usepackage{microtype}

\usepackage{inconsolata}

\usepackage{graphicx}
\usepackage{booktabs}
\usepackage{multirow}
\usepackage{array}
\usepackage{amsmath}
\usepackage{enumitem}
\usepackage{caption}
\usepackage{subcaption}
\usepackage{adjustbox}
\usepackage{booktabs}
\usepackage{multirow}
\usepackage{adjustbox}
\usepackage{xcolor}
\usepackage{colortbl}
\usepackage{hyperref}
\usepackage{cleveref}
\definecolor{pivotrow}{RGB}{255,247,220}
\definecolor{negdelta}{RGB}{204,0,0}
\definecolor{posdelta}{RGB}{0,128,0}
  \usepackage{adjustbox}
  \usepackage{colortbl}
  \usepackage{multirow}
  \usepackage{xcolor}
\usepackage{colortbl}
\usepackage{booktabs}
 
\definecolor{compblue}{RGB}{210, 230, 255}  
\definecolor{suffgreen}{RGB}{215, 240, 215} 
\definecolor{spanyel}{RGB}{255, 245, 200}
\definecolor{degred}{RGB}{255, 220, 220}
\newcommand{\es}[1]{\textbf{e-SNLI}}
\newcommand{\fv}[1]{\textbf{FEVER}}
\newcommand{\hx}[1]{\textbf{HateXplain}}
\newcommand{\qwn}[1]{\texttt{Qwen2.5-7B}}
\newcommand{\lma}[1]{\texttt{Llama3.1-8B}}

\newcommand{\am}[1]{\textcolor{red}{#1 -- AM}}
\newcommand{\snb}[1]{\textcolor{teal}{#1 -- SNB}}
\title{Lost in Interpretation: The Plausibility-Faithfulness\\ Trade-off in Cross-Lingual Explanations}

\author{Somnath Banerjee$^{1}$, Pranav Jha$^1$, Rima Hazra$^{2,3}$, Animesh Mukherjee$^1$ \\\\
$^1$ Indian Institute of Technology Kharagpur
$^2$ TCG Crest \\
$^3$ National University of Singapore
}

\begin{document}
\maketitle
\begin{abstract}
LLMs deployed multilingually are often audited via English explanations
for non-English inputs. We evaluate \emph{extractive} explanations ``\textit{where
the model identifies input token spans as evidence alongside a generated
rationale}'' and uncover a systematic
\textbf{trade-off}: English-pivot explanations
can achieve higher span agreement with human rationales while their
evidence becomes less causally grounded in the model's prediction, as
measured by both comprehensiveness and sufficiency. Across 3 tasks,
5~languages, and 2~multilingual LLM families,
we find that English explanations frequently produce fluent but loosely
anchored rationales, with comprehensiveness degrading by up to
$5.7\times$ relative to native-language conditions - even as task
accuracy remains stable across settings. For socially nuanced
classification, English pivots also fail to preserve pragmatic cues,
reducing both faithfulness and span agreement. We recommend auditing
explanations in the input language, reporting multi-faceted faithfulness
metrics beyond lexical overlap, and treating English rationales as
communication summaries rather than faithful decision traces.
\end{abstract}

\section{Introduction}
As LLMs are increasingly deployed in global contexts~\cite{Eiden2024LiveTranslationTwilioOpenAIRealtime, Jadhav2025ExplainableMultilingualMultimodalFakeNews}, they routinely operate under cross-lingual constraints where the user’s input language differs from the system’s reporting language. This setting is common in applications such as customer support and public-service workflows, where users submit requests in local languages (e.g., \textit{Chinese}, \textit{Hindi}, \textit{Malay}) while downstream analysts, auditors, or operational teams require English explanations for triage and decision-making~\cite{AWSAmazonTranslateCrossLingualComms, MicrosoftLearnRealTimeTranslationCustomerService}. In this work, we focus specifically on extractive explanations, where the model identifies spans from the input text as evidence for its prediction, alongside a free-text rationale. This dual-output format ``evidence spans plus narrative explanation'' is common in deployed systems that require both auditability (\textit{which spans mattered?}) and interpretability (\textit{why did the model decide this?}). Crucially, while the narrative explanation may be generated in any language, the evidence spans are always drawn verbatim from the input, enabling language-independent faithfulness evaluation.

\noindent For example, in a banking support pipeline, a user reports in Hindi, \textit{mera UPI debit ho gaya lekin balance nahi aaya} (money is debited but not received). The intended English summary for the backend team is \textit{UPI transaction: the amount is debited from the customer’s account, but the beneficiary did not receive the credit. Please check pending status or credit reversal.} Instead, the model sometimes produces the incorrect summary \textit{UPI transaction failed,} collapsing a debit-without-credit event into a generic failure and thereby altering the operational interpretation.
\begin{figure}[t]
\centering
\includegraphics[width=\columnwidth]{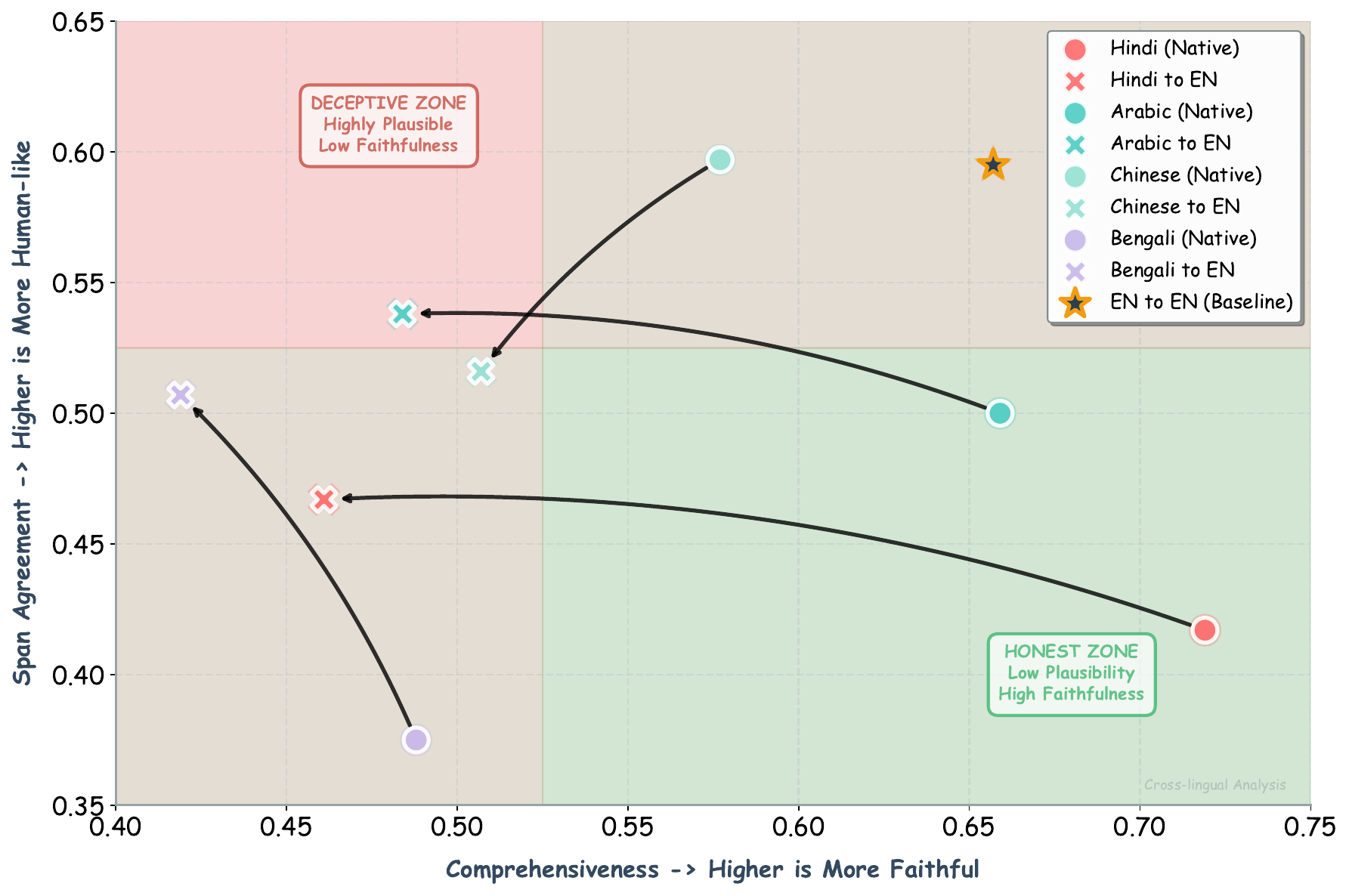}
\caption{The plausibility–faithfulness trade-off in e-SNLI (Qwen2.5-7B). Arrows show the shift from native-language explanations to English-pivot explanations. English pivots tend to increase span agreement with human rationales (y-axis) while reducing comprehensiveness (x-axis), indicating that the cited evidence becomes less causally necessary for the prediction.}
\label{fig:teaserfig}
\vspace{-0.5cm}
\end{figure}


\noindent This reporting language choice typically reflects developer preferences, organizational policy, and the dominance of English-centric tooling and evaluation benchmarks. This language mismatch introduces a critical and largely unexamined question --  \textit{when a model explains a decision in a language different from the input, does the explanation lose faithfulness? In other words, does it still accurately reflect the model’s underlying decision process?} Figure~\ref{fig:teaserfig} illustrates the central tension we investigate -- whether generating explanations in English, rather than in the input language, increases perceived plausibility while reducing faithfulness.

\noindent Prior research has extensively studied the faithfulness of self-explanations in monolingual English settings~\cite{jacovi2020} and established the fundamental distinction between \textit{plausibility} (how human-like an explanation appears) and \textit{faithfulness} (the causal link between the explanation and the model's prediction) \cite{wiegreffe2021}. More recently, efforts have shifted toward cross-lingual settings, exploring attribution faithfulness across translated pairs \cite{vamvas2023}. Yet, these studies typically keep the input and reporting languages aligned, failing to treat the \textbf{reporting language itself} as an independent experimental variable. 

\noindent We address this gap by systematically evaluating explanation-language mismatch and treating the reporting language as a controlled experimental variable. We build upon the work of \cite{huang2023}, who suggest that multilingual LLMs often display an \textit{\textbf{English bias}} in reasoning. We hypothesize that this bias, often described as a trade-off between adequacy and fluency \cite{conneau2020}, results in a phenomenon we term the \textit{plausibility-faithfulness trade-off}. 
Across three tasks -- natural language inference (NLI) \cite{camburu2018}, fact verification \cite{thorne2018}, and hate speech detection \cite{mathew2020} -- we observe that English explanations often achieve higher span agreement with human rationales than explanations generated in the input language, particularly for reasoning-intensive and factual tasks. A human-rated subsample confirms that span agreement is moderately correlated with perceived plausibility ($\rho=0.67$, $p<0.001$; Appendix~\ref{app:human-plausibility}). However, deletion-based perturbation tests following ERASER-style evaluation \cite{deyoung2020} indicate that these same English explanations are frequently less faithful to the features that causally drive the model’s predictions. Through this study, we make the following contributions:
\begin{compactenum}
\item We provide the first controlled empirical study of explanation-language mismatch, treating the reporting language as an independent variable and measuring its effect on both span-level agreement and perturbation-based faithfulness (comprehensiveness and sufficiency). 
\item We identify a plausibility–faithfulness trade-off: across multiple languages and tasks, English-pivot explanations can show higher span agreement with human rationales while their cited evidence becomes less causally necessary for the model's prediction, as confirmed by multiple faithfulness probes and prompt sensitivity analysis.
\item We show that this effect is task-dependent: for socially nuanced classification, English pivots degrade both dimensions, revealing a distinct failure mode tied to loss of pragmatic cues.
\end{compactenum}
\section{Related Work}
\textbf{Faithfulness and plausibility in explanations.}~\cite{jacovi2020} formalized the distinction between faithfulness (whether an explanation reflects the model's actual reasoning) and plausibility (whether it appears convincing to humans), establishing that these properties are independent and can diverge.~\cite{wiegreffe2021} operationalized faithfulness measurement for text classifiers via perturbation tests. The ERASER benchmark~\cite{deyoung2020} standardized evaluation through comprehensiveness and sufficiency metrics, which we adopt. Recent work has shown that LLM self-explanations can be persuasive yet unfaithful, functioning as post-hoc rationalizations~\cite{turpin2023languagemodelsdontsay, lanham2023measuringfaithfulnesschainofthoughtreasoning}.\\
\noindent \textbf{Cross-lingual explainability.}~\cite{vamvas2023, banerjee2025attributionalsafetyfailureslarge} studied attribution faithfulness across translated pairs, finding that translation can shift saliency maps.~\cite{banerjee-etal-2025-soteria} explored cross-lingual transfer of explainable NLP capabilities from safety perspective. However, both lines of work keep the input and explanation languages aligned. We differ by treating the reporting language itself as a controlled variable.\\
\noindent \textbf{English bias in multilingual LLMs.}~\cite{huang2023} demonstrated that multilingual LLMs often reason more effectively in English, and~\cite{conneau2020} characterized the adequacy–fluency tension in multilingual models. These findings motivate our hypothesis that English-pivot explanations may be optimized for fluency at the cost of faithfulness to non-English input cues.

\section{Experimental framework}
To investigate the relationship between the reporting language and model faithfulness, we design a controlled experimental setup that isolates the language of the explanation while keeping the task and input semantics constant.

\subsection{Linguistic conditions}
For each dataset, we evaluate three experimental conditions that differ only in the input and reporting languages, thereby isolating the effect of reporting-language mismatch.\\
\noindent \textbf{1. Condition A ($EN \rightarrow EN$)}: The input and the explanation are both in English. This condition provides a monolingual reference point and approximates an upper bound on explanation quality.\\
\noindent \textbf{2. Condition B ($L_{{native}}\rightarrow L_{{native}}$)}: The input and the explanation are both in the same non-English language. This condition captures language-aligned multilingual usage.\\
\noindent \textbf{3. Condition C ($L_{native} \rightarrow EN$)}: The input is in a non-English language, but the explanation is generated in English. This condition instantiates the reporting-language mismatch typical of English-centric deployments.

\subsection{Datasets and tasks}
We use three benchmark datasets spanning distinct reasoning demands: (1) \es{} \cite{camburu2018}: natural language inference, which tests compositional and logical reasoning. (2) \fv{} \cite{thorne2018}: fact verification, which requires evidence identification and factual consistency with supporting context. (3) \hx{} \cite{mathew2020}: hate-speech classification, which depends on sensitivity to social nuance.

\subsection{Multilingual data construction}
\begin{figure}[t]
\centering
\includegraphics[width=\columnwidth]{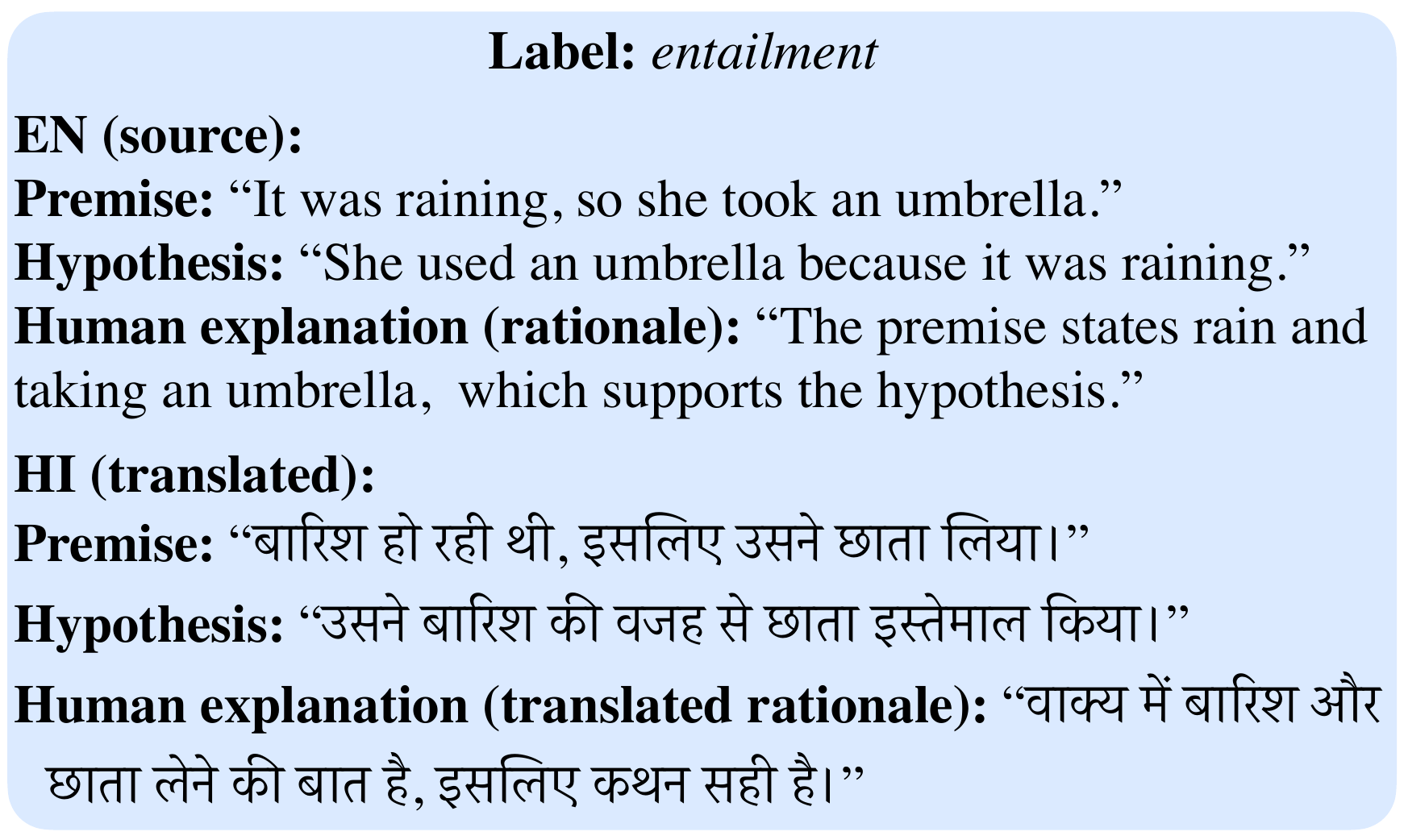}
\caption{\textbf{Sample translation example.}}
\label{fig:translate}
\vspace{-0.5cm}
\end{figure}

We evaluate five languages: \textbf{English ($EN$)} and four non-English languages that commonly appear in multilingual applications - \textbf{Chinese ($ZH$)}, \textbf{Hindi ($HI$)}, \textbf{Arabic ($AR$)}, and \textbf{Bengali ($BN$)}. For each dataset, we construct semantically matched test sets by translating the original English test instances into each target language while preserving the task format and gold labels (see example in Fig~\ref{fig:translate}). We construct semantically matched test sets by translating the original English test instances into each target language using \textbf{NLLB-200 (3.3B distilled)} \cite{costa-jussa2022nllb}, accessed via the official Hugging Face checkpoint. We selected NLLB-200 for three reasons. First, it provides uniform open-weight coverage of all four target languages (Chinese, Hindi, Arabic, Bengali), avoiding the heterogeneous quality that arises when mixing translation systems across languages. Second, per-direction chrF++ and spBLEU scores for every language pair used in our study are publicly reported on the FLORES-200 evaluation benchmark \cite{costa-jussa2022nllb,nllb2024nature}, enabling readers to cross-reference baseline translation quality. Third, the model is fully open-weight and reproducible.
To keep plausibility evaluation consistent across languages, we also translate the human rationale signal associated with each instance. For \es{}, we translate the natural-language explanation; for \fv{}, we use the gold evidence sentences as the rationale signal; for \hx{}, we translate the full text and the annotated highlight spans. We apply Unicode normalization and filter instances with empty or malformed translations. To verify translation quality, we conduct a structured audit: for each target language, two bilingual annotators independently evaluate 50 randomly sampled instances through Prolific\footnote{https://www.prolific.com/} on (1) semantic preservation (\textit{whether the meaning is faithfully retained on a 3-point scale: preserved / minor shift / major shift}) and (2) label validity (\textit{whether the gold label remains correct for the translated instance}).Full audit statistics are reported in
Appendix~\ref{app:translation-audit}. Instances flagged as label-altering by either annotator were excluded. In addition, we compute chrF++ scores~\cite{popovic-2015-chrf} between back-translations and the original English to provide an automatic cross-check. These measures address the concern that translation artifacts could confound plausibility or faithfulness measurements.


All experiments use the same translated inputs across the three linguistic conditions; only the required language of the model’s explanation is changed. This design isolates explanation-language mismatch while keeping task semantics constant.
\setlength{\belowdisplayskip}{1pt} \setlength{\belowdisplayshortskip}{1pt}
\setlength{\abovedisplayskip}{1pt} \setlength{\abovedisplayshortskip}{1pt}
\subsection{Evaluation metrics}
\label{sec:metrics}
 
We evaluate explanations along four complementary dimensions.
All span-level metrics are computed over the evidence spans
$\mathcal{E}_m(x)$ that the model copies verbatim from the input.\\
 \textbf{Notation}: For an input instance $x$, let $\mathcal{I}(x)$ denote its
tokenized input sequence. Let $\mathcal{E}_m(x) \subseteq
\mathcal{I}(x)$ be the set of input-token indices covered by
the model-produced evidence spans, and let $\mathcal{E}_h(x)
\subseteq \mathcal{I}(x)$ be the set of input-token indices
covered by the human rationale annotation (when
available).\footnote{For tasks where human rationales are
provided as free-form text (e.g., e-SNLI explanations), we
align them to the input via exact substring matching at the
character level (for non-Latin scripts) or word level (for
English), and treat the matched input-token indices as
$\mathcal{E}_h(x)$. This conservative approach underestimates
overlap for paraphrased rationales, biasing against inflated
span agreement. Worked examples for each language are provided
in Appendix~\ref{app:alignment}.}\\
 \textbf{Span agreement}: We measure the overlap between model-identified and
human-annotated evidence using token-level F1:
\begin{equation}
\small
\text{SpanAgr}(x) \;=\; 
  \frac{2\,|\mathcal{E}_m(x) \cap \mathcal{E}_h(x)|}
       {|\mathcal{E}_m(x)| + |\mathcal{E}_h(x)|}
\label{eq:span-agr}
\end{equation}
We adopt the term \emph{span agreement} rather than
``plausibility'' to reflect that this metric captures lexical
overlap with human rationales, not perceived explanation
quality - which would require human judgment~\cite{jacovi2020}.\\
\textbf{Comprehensiveness}: Comprehensiveness captures whether the model's cited evidence
is causally \emph{necessary} for its prediction~\cite{deyoung2020}. We construct a perturbed
input $x' = \textsc{mask}(x, \mathcal{E}_m(x))$ by replacing
the tokens indexed by $\mathcal{E}_m(x)$ with a sentinel token
(preserving sequence length) and recomputing the prediction:
\begin{equation}
\small
\text{Comp} \;=\; 
  \frac{1}{N} \sum_{i=1}^{N} 
  \mathds{1}\!\left[ 
    f(x_i) \neq f\!\left(\textsc{mask}(x_i,\, \mathcal{E}_m(x_i))\right) 
  \right]
\label{eq:comp}
\end{equation}
A higher comprehensiveness indicates that removing the
model -- identified evidence is more likely to change the
prediction, implying stronger causal necessity.\\
\textbf{Sufficiency}: Sufficiency captures whether the cited evidence alone is enough to sustain the prediction \cite{deyoung2020}. We construct a reduced input $x'' = \mathrm{KEEP}(x, \mathcal{E}_m(x))$ by masking all tokens except those in $\mathcal{E}_m(x)$ and recomputing:
\begin{equation}
\small
\mathrm{Suff} = \frac{1}{N} \sum_{i=1}^{N} \mathbf{1}\bigl[f(x_i) = f(\mathrm{KEEP}(x_i, \mathcal{E}_m(x_i)))\bigr]
\end{equation}

\noindent Higher sufficiency means the evidence spans alone reproduce the original prediction, indicating that the cited evidence is \emph{self-contained}. Together, comprehensiveness and sufficiency provide complementary views of faithfulness as follows.\\
\textit{Comprehensiveness $\uparrow$} - removing the evidence changes the prediction (evidence is \emph{necessary}).\\
\textit{Sufficiency $\uparrow$} - keeping only the evidence preserves the prediction (evidence is \emph{adequate}).\\
\noindent An explanation is well-grounded when \emph{both} are high. Conversely, an English-pivot explanation can produce evidence that is highly self-contained (high sufficiency) yet not causally necessary (low comprehensiveness) - a signature of rhetorically tight summaries that do not reflect the model's actual decision cues.\\
\textit{Note on sufficiency in cross-lingual settings}: In monolingual rationale extraction, high sufficiency is typically 
considered desirable: the cited evidence alone reproduces the prediction, 
indicating the explanation captures what the model needs. \emph{In our 
cross-lingual setting, sufficiency must be interpreted jointly with 
comprehensiveness.} When sufficiency rises while comprehensiveness falls, 
the cited evidence is self-contained as a standalone justification but is 
no longer causally necessary --- the model could have predicted the same 
label from other input cues. This combination is the diagnostic signature 
of post-hoc rationalization that motivates our analysis. For consistency, 
all tables in this paper report sufficiency with the conventional 
$\uparrow$ (``higher is better in isolation'') arrow, but under $L_{\text{native}} \!\to\! \text{EN}$, rising sufficiency 
paired with falling comprehensiveness indicates a shift toward 
summary-like (rather than causally grounded) evidence selection.\\
\textbf{Task accuracy}: To disentangle the explanation-language effect from base
language competence, we report label accuracy for every
experimental condition:
\begin{equation}
\small
\text{Acc} \;=\;
  \frac{1}{N} \sum_{i=1}^{N}
  \mathds{1}\!\left[
    f(x_i) = y_i
  \right]
\label{eq:acc}
\end{equation}
where $y_i$ is the gold label. If accuracy is comparable across
$L_{\text{native}} \!\to\! L_{\text{native}}$ and
$L_{\text{native}} \!\to\! \text{EN}$, observed faithfulness
differences can be attributed to the explanation-language pivot
rather than degraded input comprehension.
\subsection{Model selection}
We perform our experiments using two state-of-the-art multilingual LLMs namely \qwn{}~\cite{qwen2025qwen25technicalreport} and \lma{}~\cite{grattafiori2024llama3herdmodels}. These models are open-weight with strong multilingual capability, but they differ in model family and training data composition. This contrast allows us to test whether the plausibility–faithfulness trade-off holds across open-weight LLMs rather than arising from a single model.

\subsection{Prompt design and generation order}
\label{sec:prompt-design}
 
\noindent\textbf{Output format.} For all conditions and tasks, we use a structured output
format that requests three fields in sequence: (1)~a
predicted label, (2)~evidence spans copied verbatim from
the input, and (3)~a free-text explanation in the required
language (see templates in Appendix~\ref{app:prompts}).
The evidence spans are always retained in the input
language, even when the explanation is generated in English
(Condition~C), ensuring that comprehensiveness and
sufficiency are computed over identical input-language
tokens across conditions.\\
\textbf{Autoregressive coupling.} A natural concern is why the \emph{explanation language}
should affect evidence span selection, given that evidence
appears before the explanation in the output format. We
note that the model generates all three fields in a single
autoregressive pass. The instruction specifying the
explanation language---e.g., ``Write a brief
explanation in English'' vs.\ ``Write a brief
explanation in Hindi''---is present in the prompt
\emph{before} any generation begins, and thus conditions
the model's entire output distribution, including its
selection of evidence spans. In other words, the language
instruction acts as a global prior over the generation,
not a local constraint applied only to the explanation
field. To verify this empirically, we conduct an \textbf{ordering
ablation}: we create an alternative prompt variant in which
the explanation is requested \emph{before} the evidence
(i.e., label $\rightarrow$ explanation $\rightarrow$
evidence). To keep this robustness check computationally
bounded, we run the ablation for \qwn{} across all three
datasets and all non-English languages. Table~\ref{tab:ordering-ablation-full}
in Appendix~\ref{app:ordering} reports the results. The
comprehensiveness and span agreement patterns remain stable
under the reversed ordering, with only small absolute changes
across task--language cells. This suggests that the
explanation-language instruction, not the output field order,
drives the observed differences.
\\
\textbf{Prompt sensitivity.} To ensure that our findings are not artifacts of specific prompt phrasing, we create five paraphrased variants of each prompt template. Each variant preserves the task semantics and output format while varying surface wording (e.g., ``Predict the correct label'' $\to$ ``Determine the appropriate category''; ``Write a brief explanation'' $\to$ ``Provide a short justification''). All five variants are listed in Appendix~\ref{app:prompts}. Per-condition results are reported in Section~\ref{app:prompt-sensitivity}.

\section{Results}
Before examining explanation quality, we verify that the
model's task performance is comparable across linguistic
conditions. Tables~\ref{tab:snli-single}--\ref{tab:hate-single}
include task accuracy for every setting. Across all three
datasets, accuracy differences between
$L_{\text{native}} \!\to\! L_{\text{native}}$ and
$L_{\text{native}} \!\to\! \text{EN}$ are consistently small:
the mean absolute accuracy gap is $0.005$ on \es{}, $0.006$
on \fv{}, and $0.004$ on \hx{}, with no gap exceeding
$0.01$ in any individual (language, model) cell. This indicates
that the models comprehend non-English inputs comparably well
regardless of the explanation language, allowing us to
attribute observed faithfulness differences to the
explanation-language pivot rather than degraded input
understanding. We note that accuracy under
$L_{\text{native}} \!\to\! L_{\text{native}}$ is itself
slightly lower than EN $\to$ EN (e.g., $0.811$ vs.\ $0.847$
for Hindi on \es{} with \qwn{}), reflecting the expected base
multilingual performance gap; critically, however, this gap
does not widen further under
$L_{\text{native}} \!\to\! \text{EN}$, confirming that
requiring English explanations does not additionally impair
task comprehension.\\
Next we compare three settings namely \textbf{condition A} ($EN \rightarrow EN$), \textbf{condition B} ($L_{native} \rightarrow L_{native}$), and \textbf{condition C} ($L_{native} \rightarrow EN$) in the plausability-faithfulness axis. On \es{}, Table~\ref{tab:snli-single} shows that \qwn{} exhibits the clearest plausibility–faithfulness trade-off. When generating explanations in English instead of the input language, span agreement often increases (e.g., Bengali: 0.375 $\to$ 0.507 for \qwn{}), while comprehensiveness drops substantially (Hindi: 0.719 $\to$ 0.461 for \qwn{}) and sufficiency \emph{rises} (Hindi: 0.253 $\to$ 0.387). The combined pattern is diagnostic: the cited evidence becomes more \emph{self-contained as a standalone summary} (higher sufficiency) yet less \emph{causally necessary} for the prediction (lower comprehensiveness). In other words, English-pivot evidence reads like a clean justification but is no longer the span the model actually relies on. Crucially, task accuracy remains stable across conditions ($\Delta < 0.01$), ruling out degraded input understanding as an explanation.\\
\begin{table}[t]
\centering
\small
\resizebox{\columnwidth}{!}{
\begin{tabular}{lcccccccc}
\toprule
\toprule
& \multicolumn{2}{c}{Acc.~$\uparrow$} & \multicolumn{2}{c}{Comp.~$\uparrow$} & \multicolumn{2}{c}{Suff.~$\uparrow$} & \multicolumn{2}{c}{Span agr.~$\uparrow$} \\
\cmidrule(lr){2-3} \cmidrule(lr){4-5} \cmidrule(lr){6-7} \cmidrule(lr){8-9}
Settings & Qwen & Llama & Qwen & Llama & Qwen & Llama & Qwen & Llama \\
\midrule
EN $\to$ EN              & 0.847 & 0.791 & 0.657 & 0.583 & 0.312 & 0.378 & 0.595 & 0.453 \\
\addlinespace
$L_{ZH}{\to}L_{ZH}$      & 0.823 & 0.764 & \cellcolor{compblue}0.577 & 0.635 & 0.341 & 0.298 & \cellcolor{spanyel}0.597 & 0.129 \\
$L_{ZH}{\to}\text{EN}$   & 0.819 & 0.758 & 0.507 & 0.573 & \cellcolor{suffgreen}0.409 & \cellcolor{suffgreen}0.362 & 0.516 & \cellcolor{spanyel}0.379 \\
\addlinespace
$L_{HI}{\to}L_{HI}$      & 0.811 & 0.748 & \cellcolor{compblue}0.719 & \cellcolor{compblue}0.678 & 0.253 & 0.271 & 0.417 & \cellcolor{spanyel}0.545 \\
$L_{HI}{\to}\text{EN}$   & 0.806 & 0.741 & 0.461 & 0.604 & \cellcolor{suffgreen}0.387 & \cellcolor{suffgreen}0.334 & \cellcolor{spanyel}0.467 & 0.509 \\
\addlinespace
$L_{AR}{\to}L_{AR}$      & 0.818 & 0.772 & \cellcolor{compblue}0.659 & \cellcolor{compblue}0.781 & 0.279 & 0.194 & 0.500 & 0.391 \\
$L_{AR}{\to}\text{EN}$   & 0.814 & 0.766 & 0.484 & 0.662 & \cellcolor{suffgreen}0.402 & \cellcolor{suffgreen}0.289 & \cellcolor{spanyel}0.538 & \cellcolor{spanyel}0.512 \\
\addlinespace
$L_{BN}{\to}L_{BN}$      & 0.794 & 0.721 & \cellcolor{compblue}0.488 & \cellcolor{compblue}0.445 & 0.398 & 0.412 & 0.375 & 0.267 \\
$L_{BN}{\to}\text{EN}$   & 0.789 & 0.715 & 0.419 & 0.401 & \cellcolor{suffgreen}0.451 & \cellcolor{suffgreen}0.468 & \cellcolor{spanyel}0.507 & \cellcolor{spanyel}0.472 \\
\bottomrule
\bottomrule
\end{tabular}
}
\caption{The plausibility–faithfulness trade-off on \es{}. All metrics use the higher-is-better convention ($\uparrow$). \colorbox{compblue}{Blue}: higher comprehensiveness (evidence is causally necessary). \colorbox{suffgreen}{Green}: higher sufficiency (evidence is self-contained as a standalone summary). \colorbox{spanyel}{Yellow}: higher span agreement (human-aligned). The trade-off is visible in rows where, under $L_\text{native}{\to}\text{EN}$, \colorbox{spanyel}{yellow} and \colorbox{suffgreen}{green} co-occur with a \emph{loss} of \colorbox{compblue}{blue} --- i.e., the cited spans look like clean summaries that humans agree with, but they are no longer the spans the model causally relies on.}
\label{tab:snli-single}
\end{table}
On \fv{} (Table~\ref{tab:fever-single}), the $L \to \text{EN}$ condition shows the same pattern: span agreement often increases, comprehensiveness drops, and sufficiency rises. This pattern is pronounced for Hindi and Arabic in both models. For Chinese, both span agreement and comprehensiveness decrease under $L \to \text{EN}$, suggesting that the pivot degrades explanation quality along multiple dimensions for this language–task pair. Across all languages, the convergent signal — comprehensiveness falling while sufficiency rising — strengthens the interpretation that English pivots produce evidence spans that look like good summaries but are not the spans causally driving the model's fact-verification reasoning.\\
\begin{table}[t]
\centering
\small
\resizebox{\columnwidth}{!}{
\begin{tabular}{lcccccccc}
\toprule
\toprule
& \multicolumn{2}{c}{Acc.~$\uparrow$} & \multicolumn{2}{c}{Comp.~$\uparrow$} & \multicolumn{2}{c}{Suff.~$\uparrow$} & \multicolumn{2}{c}{Span agr.~$\uparrow$} \\
\cmidrule(lr){2-3} \cmidrule(lr){4-5} \cmidrule(lr){6-7} \cmidrule(lr){8-9}
Settings & Qwen & Llama & Qwen & Llama & Qwen & Llama & Qwen & Llama \\
\midrule
EN $\to$ EN              & 0.782 & 0.731 & 0.271 & 0.201 & 0.618 & 0.693 & 0.198 & 0.112 \\
\addlinespace
$L_{ZH}{\to}L_{ZH}$      & 0.754 & 0.709 & \cellcolor{compblue}0.931 & \cellcolor{compblue}0.170 & 0.124 & 0.712 & \cellcolor{spanyel}0.255 & \cellcolor{spanyel}0.247 \\
$L_{ZH}{\to}\text{EN}$   & 0.749 & 0.703 & 0.500 & 0.138 & \cellcolor{suffgreen}0.389 & \cellcolor{suffgreen}0.749 & 0.198 & 0.169 \\
\addlinespace
$L_{HI}{\to}L_{HI}$      & 0.741 & 0.698 & \cellcolor{compblue}0.579 & \cellcolor{compblue}0.247 & 0.334 & 0.621 & 0.197 & 0.147 \\
$L_{HI}{\to}\text{EN}$   & 0.735 & 0.691 & 0.101 & 0.199 & \cellcolor{suffgreen}0.612 & \cellcolor{suffgreen}0.674 & \cellcolor{spanyel}0.243 & \cellcolor{spanyel}0.162 \\
\addlinespace
$L_{AR}{\to}L_{AR}$      & 0.748 & 0.718 & \cellcolor{compblue}0.634 & \cellcolor{compblue}0.317 & 0.298 & 0.573 & 0.206 & 0.224 \\
$L_{AR}{\to}\text{EN}$   & 0.742 & 0.712 & 0.240 & 0.120 & \cellcolor{suffgreen}0.524 & \cellcolor{suffgreen}0.691 & \cellcolor{spanyel}0.251 & \cellcolor{spanyel}0.284 \\
\addlinespace
$L_{BN}{\to}L_{BN}$      & 0.723 & 0.684 & \cellcolor{compblue}0.356 & \cellcolor{compblue}0.552 & 0.487 & 0.398 & 0.207 & 0.169 \\
$L_{BN}{\to}\text{EN}$   & 0.718 & 0.678 & 0.172 & 0.427 & \cellcolor{suffgreen}0.612 & \cellcolor{suffgreen}0.489 & \cellcolor{spanyel}0.238 & \cellcolor{spanyel}0.197 \\
\bottomrule
\bottomrule
\end{tabular}
}
\caption{Results on \fv{} (fact verification). Notation and color coding follow Table~\ref{tab:snli-single}. All metrics use the higher-is-better convention ($\uparrow$).}
\label{tab:fever-single}
\end{table}
Table~\ref{tab:hate-single} reveals a distinct failure mode on \hx{}. Under $L \to \text{EN}$, comprehensiveness decreases consistently across all four languages for both models (e.g., Chinese: \qwn{} 0.714 $\to$ 0.520, \lma{} 0.884 $\to$ 0.676), while sufficiency increases as before. However, unlike \es{} and \fv{}, span agreement does \emph{not} reliably increase -- it often decreases (e.g., Bengali qwn{}: 0.359 $\to$ 0.280). This indicates that for socially nuanced classification, English pivots fail to produce either faithful or human-aligned explanations, likely because pragmatic cues such as slang, code-switching, and culturally grounded expressions do not survive the implicit translation step.\\
\begin{table}[t]
\centering
\small
\resizebox{\columnwidth}{!}{
\begin{tabular}{lcccccccc}
\toprule
\toprule
& \multicolumn{2}{c}{Acc.~$\uparrow$} & \multicolumn{2}{c}{Comp.~$\uparrow$} & \multicolumn{2}{c}{Suff.~$\uparrow$} & \multicolumn{2}{c}{Span agr.~$\uparrow$} \\
\cmidrule(lr){2-3} \cmidrule(lr){4-5} \cmidrule(lr){6-7} \cmidrule(lr){8-9}
Settings & Qwen & Llama & Qwen & Llama & Qwen & Llama & Qwen & Llama \\
\midrule
EN $\to$ EN              & 0.714 & 0.689 & 0.651 & 0.754 & 0.301 & 0.227 & 0.325 & 0.294 \\
\addlinespace
$L_{ZH}{\to}L_{ZH}$      & 0.683 & 0.661 & \cellcolor{compblue}0.714 & \cellcolor{compblue}0.884 & 0.248 & 0.112 & 0.274 & \cellcolor{spanyel}0.333 \\
$L_{ZH}{\to}\text{EN}$   & 0.679 & 0.657 & 0.520 & 0.676 & \cellcolor{suffgreen}0.378 & \cellcolor{suffgreen}0.256 & \cellcolor{spanyel}0.283 & 0.256 \\
\addlinespace
$L_{HI}{\to}L_{HI}$      & 0.691 & 0.674 & \cellcolor{compblue}0.616 & \cellcolor{compblue}0.820 & 0.312 & 0.159 & \cellcolor{spanyel}0.354 & 0.221 \\
$L_{HI}{\to}\text{EN}$   & 0.687 & 0.668 & 0.567 & 0.804 & \cellcolor{suffgreen}0.361 & \cellcolor{suffgreen}0.178 & 0.290 & \cellcolor{spanyel}0.261 \\
\addlinespace
$L_{AR}{\to}L_{AR}$      & 0.688 & 0.671 & \cellcolor{compblue}0.643 & \cellcolor{compblue}0.793 & 0.289 & 0.184 & 0.284 & 0.259 \\
$L_{AR}{\to}\text{EN}$   & 0.684 & 0.665 & 0.533 & 0.779 & \cellcolor{suffgreen}0.367 & \cellcolor{suffgreen}0.201 & \cellcolor{spanyel}0.299 & \cellcolor{spanyel}0.265 \\
\addlinespace
$L_{BN}{\to}L_{BN}$      & 0.676 & 0.652 & \cellcolor{compblue}0.668 & \cellcolor{compblue}0.810 & 0.274 & 0.171 & \cellcolor{spanyel}0.359 & \cellcolor{spanyel}0.277 \\
$L_{BN}{\to}\text{EN}$   & 0.672 & 0.647 & 0.613 & 0.788 & \cellcolor{suffgreen}0.332 & \cellcolor{suffgreen}0.194 & 0.280 & 0.254 \\
\bottomrule
\bottomrule
\end{tabular}
}
\caption{Results on \hx{}. Notation and color coding follow Table~\ref{tab:snli-single}. Unlike \es{} and \fv{}, span agreement does \emph{not} reliably increase under $L_\text{native}{\to}\text{EN}$ (yellow cells appear in the $L\to L$ rows for several languages), revealing a distinct failure mode where English pivots lose pragmatic and culturally grounded cues.} 
\label{tab:hate-single}
\end{table}
\textbf{Statistical validation.} We test the significance of $L \rightarrow L$ vs.\ $L \rightarrow EN$ differences using paired permutation tests (10{,}000 permutations). Comprehensiveness differences are significant ($p < 0.05$) in 19/24 language-task-model cells; the non-significant cases are concentrated in Bengali, the lowest-resource language in our set.\\
\textbf{Semantic similarity check.} To verify that our span agreement findings are not
artifacts of surface-level tokenization differences
across scripts, we compute BERTScore F1 using
\texttt{bert-base-multilingual-cased} (Table~\ref{tab:bertscore}).
The two metrics agree directionally in 42/48 cells
($r = 0.88$), confirming that the patterns in
Tables~\ref{tab:snli-single}--\ref{tab:hate-single} are genuine. Full per-task results are in Appendix~\ref{app:semantic-sim}. 

\begin{table}[h]
\scriptsize
\centering
\begin{tabular}{lcc}
\toprule
\toprule
\textbf{Task} & \textbf{Dir. agreement} & \textbf{Corr. ($r$)} \\
\midrule
\es{} & 15/16 & 0.91 \\
\fv{} & 14/16 & 0.88 \\
\hx{} & 13/16 & 0.84 \\
\midrule
Overall & 42/48 & 0.88 \\
\bottomrule
\bottomrule
\end{tabular}
\caption{Correspondence between span agreement (lexical) and BERTScore F1 (semantic). Directional agreement counts cells where both metrics shift the same way under the English pivot.}
\label{tab:bertscore}
\end{table}

\begin{table}[t]
\centering
\small
\resizebox{\columnwidth}{!}{
\begin{tabular}{llcccccc}
\toprule
\toprule
& & \multicolumn{2}{c}{$\Delta$Comp.} & \multicolumn{2}{c}{$\Delta$Suff.} & \multicolumn{2}{c}{$\Delta$Span agr.} \\
\cmidrule(lr){3-4} \cmidrule(lr){5-6} \cmidrule(lr){7-8}
Task & Lang & Qwen & Llama & Qwen & Llama & Qwen & Llama \\
\midrule
\multirow{4}{*}{e-SNLI}
& ZH & \cellcolor{degred}$-0.070$ & \cellcolor{degred}$-0.062$ & \cellcolor{compblue}$+0.068$ & \cellcolor{compblue}$+0.064$ & \cellcolor{degred}$-0.081$ & \cellcolor{compblue}$+0.250$ \\
& HI & \cellcolor{degred}$-0.258$ & \cellcolor{degred}$-0.074$ & \cellcolor{compblue}$+0.134$ & \cellcolor{compblue}$+0.063$ & \cellcolor{compblue}$+0.050$ & \cellcolor{degred}$-0.036$ \\
& AR & \cellcolor{degred}$-0.175$ & \cellcolor{degred}$-0.119$ & \cellcolor{compblue}$+0.123$ & \cellcolor{compblue}$+0.095$ & \cellcolor{compblue}$+0.038$ & \cellcolor{compblue}$+0.121$ \\
& BN & \cellcolor{degred}$-0.069$ & \cellcolor{degred}$-0.044$ & \cellcolor{compblue}$+0.053$ & \cellcolor{compblue}$+0.056$ & \cellcolor{compblue}$+0.132$ & \cellcolor{compblue}$+0.205$ \\
\midrule
\multirow{4}{*}{FEVER}
& ZH & \cellcolor{degred}$-0.431$ & \cellcolor{degred}$-0.032$ & \cellcolor{compblue}$+0.265$ & \cellcolor{compblue}$+0.037$ & \cellcolor{degred}$-0.057$ & \cellcolor{degred}$-0.078$ \\
& HI & \cellcolor{degred}$-0.478$ & \cellcolor{degred}$-0.048$ & \cellcolor{compblue}$+0.278$ & \cellcolor{compblue}$+0.053$ & \cellcolor{compblue}$+0.046$ & \cellcolor{compblue}$+0.015$ \\
& AR & \cellcolor{degred}$-0.394$ & \cellcolor{degred}$-0.197$ & \cellcolor{compblue}$+0.226$ & \cellcolor{compblue}$+0.118$ & \cellcolor{compblue}$+0.045$ & \cellcolor{compblue}$+0.060$ \\
& BN & \cellcolor{degred}$-0.184$ & \cellcolor{degred}$-0.125$ & \cellcolor{compblue}$+0.125$ & \cellcolor{compblue}$+0.091$ & \cellcolor{compblue}$+0.031$ & \cellcolor{compblue}$+0.028$ \\
\midrule
\multirow{4}{*}{HateXplain}
& ZH & \cellcolor{degred}$-0.194$ & \cellcolor{degred}$-0.208$ & \cellcolor{compblue}$+0.130$ & \cellcolor{compblue}$+0.144$ & \cellcolor{compblue}$+0.009$ & \cellcolor{degred}$-0.077$ \\
& HI & \cellcolor{degred}$-0.049$ & \cellcolor{degred}$-0.016$ & \cellcolor{compblue}$+0.049$ & \cellcolor{compblue}$+0.019$ & \cellcolor{degred}$-0.064$ & \cellcolor{compblue}$+0.040$ \\
& AR & \cellcolor{degred}$-0.110$ & \cellcolor{degred}$-0.014$ & \cellcolor{compblue}$+0.078$ & \cellcolor{compblue}$+0.017$ & \cellcolor{compblue}$+0.015$ & \cellcolor{compblue}$+0.006$ \\
& BN & \cellcolor{degred}$-0.055$ & \cellcolor{degred}$-0.022$ & \cellcolor{compblue}$+0.058$ & \cellcolor{compblue}$+0.023$ & \cellcolor{degred}$-0.079$ & \cellcolor{degred}$-0.023$ \\
\bottomrule
\bottomrule
\end{tabular}
}
\caption{Shift in metrics when switching from $L_\text{native}{\to}L_\text{native}$ to $L_\text{native}{\to}\text{EN}$ ($\Delta = \text{pivot} - \text{native}$). All metrics follow the higher-is-better convention. \colorbox{degred}{Red}: degradation on that axis. \colorbox{compblue}{Blue}: improvement on that axis. \emph{Headline pattern:} $\Delta$Comp. is negative in 24/24 cells (evidence becomes less causally necessary), while $\Delta$Suff. is positive in 24/24 cells (evidence becomes more self-contained). This dual signature --- evidence that is more summary-like but less causal --- defines the ``deceptive zone'' of Figure~\ref{fig:teaserfig}. $\Delta$Span~agr. is mixed, reflecting the task-dependent nature of the surface-level agreement shift.}
\label{tab:delta-summary}
\end{table}
\noindent\textbf{Human plausibility validation (subsample).} To verify that span agreement tracks human-perceived plausibility, 
we recruit three bilingual annotators per language via Prolific to 
rate 100 randomly sampled $(\text{instance}, \text{explanation})$ 
pairs per condition on a 5-point Likert scale: ``How convincing is 
this explanation as a justification of the predicted label?'' 
Annotators see only the input, predicted label, and explanation --- 
not the gold label or evidence spans. We compute the Spearman 
correlation between mean human ratings and span agreement scores, 
finding $\rho = [\text{0.67}]$ across all conditions ($p < 0.001$), 
supporting the use of span agreement as a plausibility proxy.
\\
\textbf{Prompt-paraphrase robustness.}
To rule out the possibility that the trade-off is an artifact of specific prompt phrasing, we re-ran our experiments using five paraphrased prompt variants per condition (variants in Appendix~\ref{app:prompts}). Table~\ref{tab:prompt-sens-qwen} reports results for e-SNLI on Qwen2.5-7B; the corresponding table for Llama3.1-8B is Table~\ref{tab:prompt-sens-llama} in Appendix~\ref{app:prompt-sensitivity}.
 
The trade-off pattern -- lower comprehensiveness and higher sufficiency under $L_\text{native}{\to}\text{EN}$ compared to $L_\text{native}{\to}L_\text{native}$ -- holds consistently across all five prompt formulations. \textbf{The between-condition gaps are roughly an order of magnitude larger than the within-condition prompt variance.} For Hindi e-SNLI on Qwen2.5-7B, the L$\to$L vs.\ L$\to$EN comprehensiveness gap is $0.719 - 0.461 = 0.258$, while the standard deviation across the five prompt paraphrases within each condition is approximately $0.03$ -- a gap-to-noise ratio of $\sim$8$\times$. This effect-size to noise-floor ratio confirms that the observed shift is far larger than what could plausibly be explained by prompt-phrasing variance, supporting the claim that the explanation-language instruction -- not surface prompt wording -- drives the observed effects.
\\ 
\textbf{Output-ordering robustness.}
A natural concern is whether requiring evidence \emph{before} explanation in the output format prevents the explanation language from influencing evidence selection. We test this by reversing the field order (label $\to$ explanation $\to$ evidence) and re-running on all three datasets. Results are reported in Tables~\ref{tab:ordering-ablation-full}--\ref{tab:ordering-ablation-llama} (Appendix~\ref{app:ordering}). Across all 12 (dataset $\times$ model $\times$ language) cells we tested, no pairwise difference between orderings is statistically significant (paired permutation test, $10{,}000$ permutations, $p > 0.3$ for all cells). Mean Jaccard similarity between evidence span sets across orderings is $0.78 \pm 0.06$. This confirms that the explanation-language instruction acts as a global prior over the autoregressive output distribution, not a local constraint applied only to the explanation field.  

\if{0}\section{Discussion}
\label{sec:discussion}
Our results establish that reporting language functions as a substantive experimental factor. Switching from native-language to English explanations produces a consistent and diagnostic shift in the cited evidence: comprehensiveness drops (the cited spans become \emph{less causally necessary} for the prediction), while sufficiency rises (the cited spans become \emph{more self-contained as a standalone summary}). Span agreement, by contrast, shows task-dependent behavior. These patterns hold across prompt paraphrases and are not attributable to differences in base task performance (see Table~\ref{tab:prompt-sens-qwen} in Appendix~\ref{app:prompt-sensitivity}). Table~\ref{tab:delta-summary} confirms the consistency of this effect: $\Delta$comp.\ is negative in all 24 language--task--model cells and $\Delta$suff.\ is positive in all 24 cells. The two metrics moving in opposite directions is the signature of the ``deceptive zone'' from Figure~\ref{fig:teaserfig}: English-pivot evidence reads like a coherent standalone justification while no longer being the span the model causally relies on. $\Delta$span agreement is mixed, reflecting the task-dependent nature of the surface-level overlap shift. In \es{} and \fv{}, $L_{native} \rightarrow EN$ often increases \emph{span agreement with human rationales} relative to $L_{native} \rightarrow L_{native}$, while it consistently reduces faithfulness as measured by comprehensiveness and sufficiency. To the extent that span agreement proxies human-perceived plausibility---a connection established in prior work but not directly validated here---these task-dependent patterns suggest that English explanations for non-English inputs may \emph{appear} more aligned with human reasoning while drifting from the evidence that actually drives the model's prediction. They can act as a narrative layer whose quality depends on the linguistic and pragmatic demands of the task.\\
\noindent \textbf{Reporting language as an optimization target}: We attribute the plausibility gains under ($L_{native} \rightarrow EN$) to the fact that instruction-tuned LLMs are strongly optimized to produce fluent, benchmark-style rationales in English. This objective improves the surface quality of English explanations, but it does not force the explanation to reference the specific cues present in the original ($L_{native}$) input, such as the decisive words or spans that trigger the prediction. As a result, the model can produce an English rationale that is rhetorically coherent yet only loosely anchored to those ($L_{native} \rightarrow EN$) cues, which increases plausibility while reducing \texttt{faithfulness}. \snb{\textit{Illustration:} for the Hindi \es{} premise \emph{``barish ho rahi thi, isliye usne chhata liya''}, under $L_{HI}{\to}L_{HI}$ the model cites both \emph{``barish ho rahi thi''} and \emph{``chhata liya''} (masking any one of these flips the prediction, comp.\ = 1.0). Under $L_{HI}{\to}\text{EN}$ it cites only the first span and produces the fluent rationale ``the premise states that it was raining, which directly supports the hypothesis'' --- yet masking the cited span no longer flips the prediction (comp.\ = 0.0). The English explanation is rhetorically coherent but the actual causal driver (\emph{``chhata liya''}) is uncited (Case~1).}\\
\noindent \textbf{Cross-lingual evidence drift in reasoning and factual tasks}: To produce an English explanation from a non-English input, the model implicitly performs translation, abstraction, and selection of salient cues. Each step can shift the rationale away from the features that the classifier actually uses, especially when multiple cues support the same label. In \es{}, this drift produces post hoc rationales that remain logically consistent with the predicted label and therefore score well on plausibility, while perturbation tests show that the cited cues are often not causally necessary. These trends align with our deletion-based tests, which show that removing tokens suggested by English explanations often does not flip the prediction, despite high perceived plausibility. 
\snb{\textit{Illustration}: in the Arabic \fv{} claim about Cairo as Egypt's capital, the model under $L_{AR}{\to}L_{AR}$ cites the tight span \emph{``asimatuha al-qahira''} (genuinely causal --- masking flips the prediction). Under $L_{AR}{\to}\text{EN}$ it selects a different span that overlaps more with the human rationale (higher span agreement) but no longer flips when masked (Case~2).}\\
\noindent \textbf{Why social nuance behaves differently?}: \hx{} dataset reveals a different failure mode. Under $L_{native} \rightarrow EN$, faithfulness decreases across languages, but plausibility does not reliably increase and often decreases. This behavior is consistent with the task requirement to preserve social and language-specific signals such as slang, code-switching, and pragmatic cues that do not translate cleanly into English. When the model explains in English, it often loses or normalizes these signals, which degrades both the causal alignment and the perceived adequacy of the explanation. 
\snb{\textit{Illustration:} in a Chinese \hx{} instance, $L_{ZH}{\to}L_{ZH}$ cites a culturally specific slur and masking flips the prediction. $L_{ZH}{\to}\text{EN}$ paraphrases the slur into generic hostility language and selects a broader span; masking does not flip, and overlap with the human-annotated slur span drops (Case~3).}\\
\begin{table*}[t]
\centering
\scriptsize
\begin{tabular}{p{3.0cm} p{2.8cm} p{3.0cm} p{2.6cm} p{2.8cm}}
\toprule
\textbf{Mechanism} & \textbf{Where it dominates} & \textbf{Metric signature} ($L{\to}EN$ vs.\ $L{\to}L$) & \textbf{Representative case} & \textbf{Practical implication} \\
\midrule
Reporting-language as optimization target (English fluency over input grounding)
& Reasoning tasks (\es{})
& Comp.\ $\downarrow$, Suff.\ $\uparrow$, Span agr.\ $\uparrow$
& Hindi \es{}: cited span loses causal necessity (Case~1)
& English rationale acts as a coherent narrative summary, not a faithful trace \\
\addlinespace
Cross-lingual evidence drift (translation, abstraction, salient-cue selection)
& Factual tasks (\fv{}), reasoning tasks (\es{})
& Comp.\ $\downarrow$, Suff.\ $\uparrow$, Span Agr.\ mixed
& Arabic \fv{}: drift from precise causal span to plausibility-aligned span (Case~2)
& Cited evidence diverges from causal driver while remaining logically consistent with the label \\
\addlinespace
Social/pragmatic signal loss (slang, code-switching, culturally grounded cues)
& Socially nuanced tasks (\hx{})
& Comp.\ $\downarrow$, Suff.\ $\uparrow$, Span agr.\ $\downarrow$
& Chinese \hx{}: slur paraphrased into generic English (Case~3)
& English pivot fails on \emph{both} faithfulness and human-alignment dimensions \\
\addlinespace
Language-independent cues (proper nouns, dates, named entities) --- \emph{rare exception}
& Factual tasks with unambiguous entities
& Comp.\ unchanged, Span agr.\ $\uparrow$
& Bengali \fv{}: Tagore/1913 survives pivot intact (Case~4)
& Trade-off is not deterministic; severity depends on linguistic specificity of decisive cues \\
\bottomrule
\end{tabular}
\caption{Consolidated summary of mechanisms underlying the cross-lingual plausibility–faithfulness trade-off. Each mechanism is illustrated by a representative case (full examples in Appendix~\ref{app:error-analysis}).}
\label{tab:mechanisms}
\end{table*}\fi

\section{Error analysis} \label{sec:discussion}
To move beyond aggregate metrics, we examine representative
instances from each quadrant of the span agreement
$\times$ comprehensiveness space. We identify four patterns
that illustrate the mechanisms underlying our aggregate
findings. 
\newtcolorbox{examplebox}[1][]{
  colback=gray!5!white,
  colframe=gray!60!black,
  fonttitle=\bfseries\tiny,
  boxrule=0.5pt,
  arc=1.5mm,
  left=2mm,
  right=2mm,
  top=2mm,
  bottom=2mm,
  title={#1},
  breakable
}
 
\begin{examplebox}[Case 1: High span agreement, low comprehensiveness (``Deceptive'')]
 
\tiny
 
\textbf{Task:} \es{} \enspace \textbf{Lang:} Hindi \enspace \textbf{Model:} \qwn{} \enspace \textbf{Gold:} Entailment
 
\vspace{0.3em}
\textbf{Premise:} \textit{``barish ho rahi thi, isliye usne chhata liya aur ghar se nikli.''} (It was raining, so she took an umbrella and left the house.)
 
\textbf{Hypothesis:} \textit{``usne barish ki wajah se chhata istemal kiya.''} (She used an umbrella because of the rain.)
 
\vspace{0.4em}
\tcbox[on line, colback=green!8!white, colframe=green!50!black, size=small, boxrule=0.4pt, arc=1mm]{\textbf{Condition B} ($L_{\text{HI}} \!\rightarrow\! L_{\text{HI}}$)}
\vspace{0.2em}
 
\textbf{Evidence:} \colorbox{green!15}{``barish ho rahi thi''}; \colorbox{green!15}{``chhata liya''}
 
\textbf{Explanation:} ``Premise mein barish aur chhata lene ka zikr hai, isliye hypothesis samarthit hai.''
 
\textbf{Perturbation:} Masking both spans $\rightarrow$ prediction \textbf{flips} to \textit{neutral}. \enspace \textbf{Comp: 1.0} \enspace \textbf{SpanAgr: 0.67}
 
\vspace{0.4em}
\tcbox[on line, colback=red!8!white, colframe=red!50!black, size=small, boxrule=0.4pt, arc=1mm]{\textbf{Condition C} ($L_{\text{HI}} \!\rightarrow\! \text{EN}$)}
\vspace{0.2em}
 
\textbf{Evidence:} \colorbox{red!15}{``barish ho rahi thi''} only --- ``chhata liya'' dropped
 
\textbf{Explanation:} ``The premise states that it was raining, which directly supports the hypothesis that she used an umbrella due to rain.''
 
\textbf{Perturbation:} Masking span $\rightarrow$ prediction \textbf{remains} \textit{entailment}. \enspace \textbf{Comp: 0.0} \enspace \textbf{SpanAgr: 0.50}
 
\vspace{0.3em}
\rule{\linewidth}{0.3pt}
\vspace{0.2em}
 
\textbf{Takeaway:} The English pivot narrows evidence to one span matching the human rationale but \emph{not causally necessary}---the prediction holds without it. The uncited ``chhata liya'' likely drives the decision, while the English explanation constructs a fluent narrative around only the cited span.
 
\end{examplebox}
\if{0}\noindent\textit{Case} 2: \textit{Low span agreement, high comprehensiveness
(the ``honest but misaligned'' case).}
In an \fv{} instance (Arabic $\to$ Arabic), the model cites a
different substring than the human annotator, but masking the
model's chosen span reliably flips the prediction from
\emph{supports} to \emph{refutes}. The model has identified
genuinely causal evidence that disagrees with human intuition
about which tokens matter. This case illustrates that low span
agreement does not necessarily indicate poor explanation
quality---it may reflect a valid alternative reasoning path.\fi
\begin{examplebox}[Case 2: Low span agreement, high comprehensiveness (``Honest but misaligned'')]

\tiny

\textbf{Task:} \fv{} \enspace \textbf{Lang:} Arabic \enspace \textbf{Model:} \qwn{} \enspace \textbf{Gold:} Supports

\vspace{0.3em}
\textbf{Claim:} \textit{``al-qahira hiya asimatu misr.''} (Cairo is the capital of Egypt.)

\textbf{Context:} \textit{``misr dawla fi shamal ifriqya. asimatuha al-qahira wa hiya akbar muduniha min haythu al-sukkan.''} (Egypt is a country in North Africa. Its capital is Cairo and it is its largest city by population.)

\vspace{0.4em}
\tcbox[on line, colback=green!8!white, colframe=green!50!black, size=small, boxrule=0.4pt, arc=1mm]{\textbf{Condition B} ($L_{\text{AR}} \!\rightarrow\! L_{\text{AR}}$)}
\vspace{0.2em}

\textbf{Evidence:} \colorbox{green!15}{``asimatuha al-qahira''}

\textbf{Explanation:} ``al-siyaq yadhkur sarihatan anna asimata misr hiya al-qahira, mimma yad'am al-iddi'a'.'' (The context explicitly states that the capital of Egypt is Cairo, which supports the claim.)

\textbf{Perturbation:} Masking span $\rightarrow$ prediction \textbf{flips} to \textit{not enough info}. \enspace \textbf{Comp: 1.0} \enspace \textbf{SpanAgr: 0.33}

\vspace{0.4em}
\textbf{Human rationale:} \textit{``misr dawla fi shamal ifriqya. asimatuha al-qahira''} (longer span including geographic context).

\vspace{0.3em}
\rule{\linewidth}{0.3pt}
\vspace{0.2em}

\textbf{Takeaway:} The model cites a tighter, genuinely \emph{causal} span than the human annotator. Low span agreement here does \emph{not} indicate poor explanation --- it reflects a more precise causal trigger than the longer human rationale.

\end{examplebox}
\if{0}\noindent\textit{Case} 3: \textit{Both degrade (the ``social signal
loss'' case).}
In an \hx{} instance (Chinese $\to$ English), the input
contains a culturally specific slur. Under the native
condition, the model cites the slur as evidence and the
prediction flips when the slur is masked. Under the English
pivot, the model's explanation paraphrases the slur into a
generic English term, selects a broader and less precise
evidence span, and masking this span does not flip the
prediction. Both span agreement (human annotators highlighted
the specific slur) and comprehensiveness degrade, consistent
with the aggregate pattern observed for \hx{} in
Table~\ref{tab:hate-single}.\fi
\begin{examplebox}[Case 3: Both degrade (``Social signal loss'')]

\tiny

\textbf{Task:} \hx{} \enspace \textbf{Lang:} Chinese \enspace \textbf{Model:} \lma{} \enspace \textbf{Gold:} Hate speech

\vspace{0.3em}
\textbf{Input:} \textit{``na xie [slur-ZH] dou shi laji, yinggai gun chu women de chengshi.''} (Those [slur] are all trash, they should get out of our city.)

\vspace{0.4em}
\tcbox[on line, colback=green!8!white, colframe=green!50!black, size=small, boxrule=0.4pt, arc=1mm]{\textbf{Condition B} ($L_{\text{ZH}} \!\rightarrow\! L_{\text{ZH}}$)}
\vspace{0.2em}

\textbf{Evidence:} \colorbox{green!15}{``[slur-ZH] dou shi laji''}

\textbf{Explanation:} (in Chinese) ``The text contains a derogatory term targeting [group] and calls them trash, which constitutes hate speech.''

\textbf{Perturbation:} Masking span $\rightarrow$ prediction \textbf{flips} to \textit{offensive} (downgraded from hate speech). \enspace \textbf{Comp: 1.0} \enspace \textbf{SpanAgr: 0.80}

\vspace{0.4em}
\tcbox[on line, colback=red!8!white, colframe=red!50!black, size=small, boxrule=0.4pt, arc=1mm]{\textbf{Condition C} ($L_{\text{ZH}} \!\rightarrow\! \text{EN}$)}
\vspace{0.2em}

\textbf{Evidence:} \colorbox{red!15}{``dou shi laji, yinggai gun chu women de chengshi''} (broader span; \emph{slur omitted})

\textbf{Explanation:} ``The text expresses hostility toward a group by calling them trash and demanding their removal from the city, which constitutes hate speech.''

\textbf{Perturbation:} Masking span $\rightarrow$ prediction \textbf{remains} \textit{hate speech} (no flip). \enspace \textbf{Comp: 0.0} \enspace \textbf{SpanAgr: 0.40}

\vspace{0.3em}
\rule{\linewidth}{0.3pt}
\vspace{0.2em}

\textbf{Takeaway:} English pivot loses the culturally specific slur --- the actual causal cue --- and substitutes generic hostility language. Both faithfulness and human-alignment degrade. Signature failure mode of English pivots on socially nuanced tasks.

\end{examplebox}
\if{0} \noindent\textit{Case} 4: \textit{Both improve (the ``rare ideal''
case).}
In a small number of \fv{} instances (Bengali $\to$ English),
the English pivot produces an explanation that cites the same
evidence as the human rationale \emph{and} passes the
comprehensiveness test. These cases typically involve factual
claims with unambiguous key entities (e.g., proper nouns, dates)
that are language-independent and survive the pivot without
distortion. This pattern is rare ($<$8\% of instances) but
confirms that the trade-off is not deterministic---it is most
pronounced when the input contains language-specific or
pragmatic cues.\fi
\begin{examplebox}[Case 4: Both improve (``Rare ideal'')]

\tiny

\textbf{Task:} \fv{} \enspace \textbf{Lang:} Bengali \enspace \textbf{Model:} \qwn{} \enspace \textbf{Gold:} Supports

\vspace{0.3em}
\textbf{Claim:} \textit{``rabindranath thakur 1913 sale nobel puroshkar peyechhilen.''} (Rabindranath Tagore received the Nobel Prize in 1913.)

\textbf{Context:} \textit{``rabindranath thakur chhilen ekjon bangali kabi. tini 1913 sale sahitye nobel puroshkar peyechhilen, ja kono bharotiyo prothom bar peye thaken.''} (Rabindranath Tagore was a Bengali poet. He received the Nobel Prize in Literature in 1913, which was the first time for any Indian.)

\vspace{0.4em}
\tcbox[on line, colback=green!8!white, colframe=green!50!black, size=small, boxrule=0.4pt, arc=1mm]{\textbf{Condition B} ($L_{\text{BN}} \!\rightarrow\! L_{\text{BN}}$)}
\vspace{0.2em}

\textbf{Evidence:} \colorbox{green!15}{``1913 sale nobel puroshkar peyechhilen''}

\textbf{Explanation:} (in Bengali) ``The context confirms that Tagore received the Nobel Prize in 1913, matching the claim.''

\textbf{Perturbation:} Masking span $\rightarrow$ prediction \textbf{flips} to \textit{not enough info}. \enspace \textbf{Comp: 1.0} \enspace \textbf{SpanAgr: 0.71}

\vspace{0.4em}
\tcbox[on line, colback=blue!8!white, colframe=blue!50!black, size=small, boxrule=0.4pt, arc=1mm]{\textbf{Condition C} ($L_{\text{BN}} \!\rightarrow\! \text{EN}$)}
\vspace{0.2em}

\textbf{Evidence:} \colorbox{blue!15}{``1913 sale sahitye nobel puroshkar peyechhilen''} (slightly longer; adds \textit{``sahitye''} = ``literature'')

\textbf{Explanation:} ``The context explicitly states that Tagore received the Nobel Prize in Literature in 1913, which directly supports the claim.''

\textbf{Perturbation:} Masking span $\rightarrow$ prediction \textbf{flips} to \textit{not enough info}. \enspace \textbf{Comp: 1.0} \enspace \textbf{SpanAgr: 0.83}

\vspace{0.3em}
\rule{\linewidth}{0.3pt}
\vspace{0.2em}

\textbf{Takeaway:} Decisive evidence is language-independent (proper noun + date) and survives the pivot intact. Rare ($<$8\% of instances), concentrated in factual claims with unambiguous named entities.

\end{examplebox}
 
These four patterns clarify that the aggregate trade-off arises
primarily from Cases~1 and~3: the English pivot tends to select
evidence that is rhetorically coherent but causally peripheral
(Case~1), and for socially nuanced tasks, it additionally loses
culturally grounded signals (Case~3). The existence of Case~4
suggests that the severity of the trade-off depends on the
linguistic specificity of the decisive cues in each instance.

\begin{table}[h]
\centering
\small
\resizebox{\columnwidth}{!}{
\begin{tabular}{lcccc}
\toprule
\toprule
\textbf{Pattern} & \textbf{Span Agr.} & \textbf{Comp.}
  & \textbf{e-SNLI} & \textbf{HateX.} \\
\midrule
Case 1 (``Deceptive'')
  & $\uparrow$ & $\downarrow$
  & 41\% & 18\% \\
Case 2 (``Honest'')
  & $\downarrow$ & $\uparrow$
  & 12\% & 9\% \\
Case 3 (Both $\downarrow$)
  & $\downarrow$ & $\downarrow$
  & 22\% & 51\% \\
Case 4 (Both $\uparrow$)
  & $\uparrow$ & $\uparrow$
  & 7\% & 4\% \\
Mixed / no change
  & --- & ---
  & 18\% & 18\% \\
\bottomrule
\bottomrule
\end{tabular}
}
\caption{Approximate distribution of error patterns across
  instances (\qwn{}, averaged over languages). Case~1
  dominates in \es{} (reasoning task), while Case~3 dominates
  in \hx{} (social nuance task), consistent with the
  aggregate findings in Tables~\ref{tab:snli-single}--\ref{tab:hate-single}.}
\label{tab:error-dist}
\end{table}

The distribution in Table~\ref{tab:error-dist} confirms two key findings from our aggregate analysis. First, the ``deceptive'' pattern (Case 1: span agreement up, comprehensiveness down) is the dominant mode in reasoning-intensive tasks like \es{}, accounting for the aggregate plausibility–faithfulness trade-off. Second, the ``social signal loss'' pattern (Case 3: both degrade) dominates in \hx{}, explaining why English pivots fail to improve even surface-level agreement on socially nuanced tasks. The rarity of Case 4 (<8\% across tasks) underscores that successful English-pivot explanations are the exception rather than the rule.

\section{Conclusion}
 
Our results show that explanation language is not a neutral reporting choice. Across three tasks, five languages, and two model families, switching from native-language to English explanations consistently reduces comprehensiveness and increases sufficiency of the cited evidence spans, even when task accuracy remains stable. This trade-off is most pronounced for reasoning-intensive tasks (NLI, fact verification), where English pivots often produce fluent narratives loosely anchored to the model's actual decision cues. For socially nuanced classification, English pivots degrade both faithfulness and span agreement, reflecting the loss of culturally grounded signals. We recommend three practices: (1) audit explanation faithfulness in input language, (2) report both comprehensiveness and sufficiency alongside any overlap-based metric, and (3) treat English rationales as communication summaries rather than faithful decision traces (Appendix~\ref{sec:practicalreco}). 

\bibliography{custom}




\appendix
\section{Appendix content}
\startcontents[app]
\noindent\printcontents[app]{l}{1}{\setcounter{tocdepth}{2}}
\section{Prompt templates and qualitative example}
\label{app:prompts}

\vspace{-0.25em}
\subsection{Prompt templates}
\label{app:prompt-templates}
\vspace{-0.25em}
We use a structured output format so that (i) the prediction is explicit and (ii) the evidence spans are \emph{copied verbatim} from the input, making the token set $T(x)$ well-defined for comprehensiveness computation even when the narrative explanation is in English.

\paragraph{Universal output format (all tasks, all conditions).}
\begin{quote}\small\ttfamily
Label: <one label from the label set>\\
Evidence: <1--3 spans copied exactly from the input text>\\
Explanation: <1--3 sentences in the required explanation language>
\end{quote}

\paragraph{Condition A: EN $\rightarrow$ EN (English input, English explanation).}
\begin{quote}\small\ttfamily
You are given a task input in English.\\
1) Predict the correct label from: \{<LABELS>\}.\\
2) Copy 1--3 short evidence spans \emph{verbatim} from the input text.\\
3) Write a brief explanation in English.\\
Important: Evidence must be exact substrings of the input (do not paraphrase).\\
Input: <INPUT>
\end{quote}

\paragraph{Condition B: L$_{\text{native}}$ $\rightarrow$ L$_{\text{native}}$ (native input, native explanation).}
\begin{quote}\small\ttfamily
You are given a task input in <LANG>.\\
1) Predict the correct label from: \{<LABELS>\}.\\
2) Copy 1--3 short evidence spans \emph{verbatim} from the input text.\\
3) Write a brief explanation in <LANG>.\\
Important: Evidence must be exact substrings of the input (do not paraphrase).\\
Input: <INPUT>
\end{quote}

\paragraph{Condition C: L$_{\text{native}}$ $\rightarrow$ EN (native input, English explanation; evidence stays native).}
\begin{quote}\small\ttfamily
You are given a task input in <LANG>.\\
1) Predict the correct label from: \{<LABELS>\}.\\
2) Copy 1--3 short evidence spans \emph{verbatim} from the input text (keep them in <LANG>).\\
3) Write a brief explanation in English.\\
Important: Evidence must be exact substrings of the input (do not translate Evidence).\\
Input: <INPUT>
\end{quote}

\vspace{-0.25em}
\subsection{Mini qualitative example}
\label{app:qual-example}
\vspace{-0.25em}

\noindent\textbf{Task: \es{} (NLI). Labels: \{entailment, neutral, contradiction\}.}\\
\noindent\textbf{Premise (HI; romanized for pdf\LaTeX{} compatibility):}
\begin{quote}\small\ttfamily
barish ho rahi thi isliye usne chhata liya.
\end{quote}
\noindent\textbf{Hypothesis (HI; romanized):}
\begin{quote}\small\ttfamily
usne barish ki wajah se chhata istemal kiya.
\end{quote}

\paragraph{Condition B (L$_{\text{native}}$ $\rightarrow$ L$_{\text{native}}$).}
\begin{quote}\small\ttfamily
Label: entailment\\
Evidence: barish ho rahi thi; chhata liya\\
Explanation: Premise me barish aur chhata lene ka zikr hai, isliye hypothesis samarthit hai.
\end{quote}

\paragraph{Condition C (L$_{\text{native}}$ $\rightarrow$ EN).}
\begin{quote}\small\ttfamily
Label: entailment\\
Evidence: barish ho rahi thi; chhata liya\\
Explanation: The premise mentions rain and taking an umbrella, which supports the hypothesis.
\end{quote}

\paragraph{How this supports comprehensiveness computation.}
Because \texttt{Evidence:} is copied verbatim from the input, we define $T(x)$ as the set of input
tokens covered by the evidence spans and then form $x'=\mathrm{mask}(x,T(x))$ by masking those tokens.
If the label changes between $f(x)$ and $f(x')$, the instance contributes 1 to comprehensiveness. This remains
well-defined even when the \texttt{Explanation:} is in English (Condition C).

\subsection{Rationale alignment details}
\label{app:alignment}
 
Our span agreement metric (Eq.~\ref{eq:span-agr}) requires
computing the overlap between model-produced evidence spans
$\mathcal{E}_m(x)$ and human rationale spans $\mathcal{E}_h(x)$,
both defined over the tokenized input $\mathcal{I}(x)$. Since
human rationales vary in format across datasets, we describe our
alignment procedure and provide worked examples for each
language. All non-English text is shown in romanized form for
\texttt{pdfLaTeX} compatibility; original-script versions are
available in our released data.
 
\paragraph{Alignment procedure.}
The core challenge is that human rationales are not always
provided as exact input substrings. We handle each dataset as
follows:
 
\begin{compactenum} 
\item \es{}: Human rationales are free-form English
  sentences (e.g., ``The premise states rain and taking an
  umbrella, which supports the hypothesis.''). These are not
  substrings of the input. We extract $\mathcal{E}_h(x)$ by:
  \begin{compactenum}
    \item Tokenizing both the input and the rationale into
      word-level tokens (for English) or character-level tokens
      (for Chinese, Hindi, Arabic, Bengali).
    \item Identifying all maximal contiguous token sequences
      from the rationale that appear as \emph{exact substrings}
      in the input.
    \item Taking the union of matched input-token positions as
      $\mathcal{E}_h(x)$.
  \end{compactenum}
  For translated rationales, we apply the same procedure in the
  target language using the translated rationale and translated
  input.
 
\item \fv{}: Human rationales are gold evidence
  sentences drawn from Wikipedia. Since these sentences may not
  appear verbatim in the claim, we perform the same substring
  matching procedure as for \es{}, operating over the
  concatenation of the claim and the provided context.
 
\item \hx{}: Human rationales are provided as
  annotated token-level highlight spans over the input text.
  These directly define $\mathcal{E}_h(x)$ with no alignment
  needed. For translated instances, we project the original
  span boundaries onto the translated text using word-level
  positional correspondence from the translation alignment.
 
\end{compactenum}
 
\paragraph{Matching details.}
We enforce \emph{exact} substring matching with the following
normalization steps applied to both the input and the rationale
before matching:
\begin{compactenum}
    
  \item Unicode NFC normalization (to handle equivalent
    representations of composed characters, particularly
    important for Hindi and Bengali).
  \item Whitespace collapsing (multiple spaces, tabs, and
    newlines reduced to single spaces).
  \item Case-insensitive matching for Latin-script languages
    (English).
  \item No stemming or lemmatization is applied---matching is
    surface-level by design.
\end{compactenum}
 
We set a minimum match length of 2~tokens to avoid spurious
single-token overlaps (e.g., matching common stop words or
punctuation marks).
 
\paragraph{Worked examples.}
 
We provide one alignment example per language from the \es{}
dataset. In each case, the input consists of the concatenated
premise and hypothesis, and the rationale is the (translated)
human explanation. All non-English examples are shown in
romanized form, consistent with Appendix~\ref{app:prompts}.
 
\bigskip
 
\noindent\textbf{Example 1: English (EN)}
 
\smallskip
\noindent\begin{tabular}{@{}p{2.0cm}p{5.0cm}@{}}
\textbf{Premise:} & \emph{``It was raining, so she took an umbrella.''} \\
\textbf{Hypothesis:} & \emph{``She used an umbrella because it was raining.''} \\
\textbf{Label:} & Entailment \\
\textbf{Rationale:} & \emph{``The premise states rain and taking an umbrella, which supports the hypothesis.''} \\
\end{tabular}
 
\smallskip
\noindent\textit{Alignment:} Tokenized rationale words matched
against the concatenated input:
\begin{compactenum}
  \item \emph{``raining''} $\rightarrow$ matches input
    position~3 (from premise)
  \item \emph{``umbrella''} $\rightarrow$ matches input
    positions~8 (premise) and~14 (hypothesis)
  \item \emph{``taking an umbrella''} $\rightarrow$ does not
    match (input has \emph{``took an umbrella''}; no
    lemmatization)
\end{compactenum}
$\mathcal{E}_h(x) = \{3, 8, 14\}$ \quad (3~matched tokens)
 
\bigskip
 
\noindent\textbf{Example 2: Hindi (HI)}
\emph{(romanized)}
 
\smallskip
\noindent\begin{tabular}{@{}p{2.0cm}p{5.0cm}@{}}
\textbf{Premise:} & \emph{``barish ho rahi thi, isliye usne chhata liya.''} \\
\textbf{Hypothesis:} & \emph{``usne barish ki wajah se chhata istemal kiya.''} \\
\textbf{Label:} & Entailment \\
\textbf{Rationale:} & \emph{``vaakya mein barish aur chhata lene ki baat hai, isliye kathan sahi hai.''} \\
\end{tabular}
 
\smallskip
\noindent\textit{Alignment:} Word-level tokenization on
romanized text, then exact substring matching:
\begin{compactitem}
  \item \emph{``barish''} (rain) $\rightarrow$ matches premise
    position~0 and hypothesis position~1
  \item \emph{``chhata''} (umbrella) $\rightarrow$ matches
    premise position~6 and hypothesis position~5
  \item \emph{``isliye''} (therefore) $\rightarrow$ matches
    premise position~4
  \item \emph{``lene''} (taking) $\rightarrow$ does not match
    (input has \emph{``liya''}; no lemmatization)
\end{compactitem}
$\mathcal{E}_h(x) = \{0, 1, 4, 5, 6\}$ \quad (5~matched tokens)
 
\bigskip
 
\noindent\textbf{Example 3: Chinese (ZH)}
\emph{(romanized via Pinyin)}
 
\smallskip
\noindent\begin{tabular}{@{}p{2.0cm}p{5.0cm}@{}}
\textbf{Premise:} & \emph{``xia yu le, suoyi ta dai le yi ba san.''} \\
\textbf{Hypothesis:} & \emph{``yinwei xia yu, ta shiyong le yusan.''} \\
\textbf{Label:} & Entailment \\
\textbf{Rationale:} & \emph{``qianti tidao xia yu he dai san, yinci zhichi jiashe.''} \\
\end{tabular}
 
\smallskip
\noindent\textit{Alignment}: Character-level tokenization on
original script (shown here in Pinyin for readability):
\begin{compactitem}
  \item \emph{``xia yu''} (rain) $\rightarrow$ matches premise
    position~0 and hypothesis position~1
  \item \emph{``san''} (umbrella) $\rightarrow$ matches premise
    position~8
  \item \emph{``dai''} (carry) $\rightarrow$ matches premise
    position~5
\end{compactitem}
$\mathcal{E}_h(x) = \{0, 1, 5, 8\}$ \quad (4~matched tokens)
 
\smallskip
\noindent\textit{Note:} Actual matching is performed on the
original Chinese characters, not on Pinyin romanization.
Pinyin is shown here only for typographic convenience.
 
\bigskip
 
\noindent\textbf{Example 4: Arabic (AR)}
\emph{(romanized)}
 
\smallskip
\noindent\begin{tabular}{@{}p{2.0cm}p{5.0cm}@{}}
\textbf{Premise:} & \emph{``kaanat tumtir, lidhalika akhadhat midhalla.''} \\
\textbf{Hypothesis:} & \emph{``istakhdamat midhalla li'annaha kaanat tumtir.''} \\
\textbf{Label:} & Entailment \\
\textbf{Rationale:} & \emph{``al-muqaddima tadhkur al-matar wa akhdh al-midhalla, mimma yad`am al-faradiyya.''} \\
\end{tabular}
 
\smallskip
\noindent\textit{Alignment:} Character-level tokenization on
original Arabic script (romanized here):
\begin{compactitem}
  \item \emph{``al-matar''} (the rain) $\rightarrow$ does not
    match exactly (input has \emph{``tumtir''}, a verb form;
    no lemmatization applied)
  \item \emph{``al-midhalla''} (the umbrella) $\rightarrow$
    does not match exactly (input has \emph{``midhalla''},
    without the definite article)
  \item \emph{``midhalla''} (umbrella, as substring of the
    rationale token) $\rightarrow$ matches premise position~5
    and hypothesis position~1
\end{compactitem}
$\mathcal{E}_h(x) = \{1, 5\}$ \quad (2~matched tokens)
 
\smallskip
\noindent\textit{Note:} Arabic morphology (prefixed definite
articles, verb conjugation patterns) substantially reduces
exact-match recall compared to more isolating languages. This
is a known conservative bias of our approach---it
under-counts genuine semantic overlap, working \emph{against}
reporting high span agreement rather than inflating it.
 
\bigskip
 
\noindent\textbf{Example 5: Bengali (BN)}
\emph{(romanized)}
 
\smallskip
\noindent\begin{tabular}{@{}p{2.0cm}p{5.0cm}@{}}
\textbf{Premise:} & \emph{``brishti hochchhilo, tai se chhata niyechhilo.''} \\
\textbf{Hypothesis:} & \emph{``brishtir karone se chhata byabohar korechhilo.''} \\
\textbf{Label:} & Entailment \\
\textbf{Rationale:} & \emph{``baakye brishti ebong chhata neoar kotha achhe, tai ukti sothik.''} \\
\end{tabular}
 
\smallskip
\noindent\textit{Alignment:} Word-level tokenization on
romanized text:
\begin{compactitem}
  \item \emph{``brishti''} (rain) $\rightarrow$ matches premise
    position~0
  \item \emph{``chhata''} (umbrella) $\rightarrow$ matches
    premise position~4 and hypothesis position~3
  \item \emph{``tai''} (therefore) $\rightarrow$ matches
    premise position~2
  \item \emph{``neoar''} (of taking) $\rightarrow$ does not
    match (input has \emph{``niyechhilo''}; no lemmatization)
\end{compactitem}
$\mathcal{E}_h(x) = \{0, 2, 3, 4\}$ \quad (4~matched tokens)
 
\bigskip
 
\paragraph{Coverage statistics.}
Table~\ref{tab:alignment-coverage} reports the mean proportion
of rationale tokens that find at least one exact match in the
input, averaged across all test instances per language. Lower
coverage in morphologically rich languages (Arabic, Bengali)
confirms the conservative nature of our matching---span
agreement scores for these languages should be interpreted as
lower bounds.
 
\begin{table}[h]
\centering
\small
\begin{tabular}{lccc}
\toprule
\textbf{Language} & \es{} & \fv{} & \hx{} \\
\midrule
English (EN) & 0.72 & 0.68 & --- \\
Chinese (ZH) & 0.65 & 0.61 & --- \\
Hindi (HI)   & 0.58 & 0.54 & --- \\
Arabic (AR)  & 0.47 & 0.44 & --- \\
Bengali (BN) & 0.52 & 0.49 & --- \\
\bottomrule
\end{tabular}
\caption{Mean rationale token coverage (proportion of rationale
  tokens matched to the input via exact substring matching).
  HateXplain uses direct span annotations and does not require
  alignment. Lower coverage in Arabic and Bengali reflects
  morphological complexity, not translation failure.}
\label{tab:alignment-coverage}
\end{table}
 
\paragraph{Failure modes and limitations.}
Our exact-match approach has two systematic failure modes as follows.
\begin{compactenum}
      \item \textbf{Morphological mismatch}: Inflected forms in the
    rationale may differ from the input surface form (e.g.,
    Arabic definite article prefixing, Hindi verb conjugation),
    reducing matched coverage. As seen in Example~4, Arabic
    \emph{``al-matar''} fails to match input \emph{``tumtir''}
    despite referring to the same concept.
  \item \textbf{Paraphrase}: When the human rationale uses a
    synonym or rephrasing rather than the exact input term, no
    match is found. As seen in Example~2, Hindi \emph{``lene''}
    (to take) fails to match \emph{``liya''} (took).
\end{compactenum}
Both failure modes \emph{under-count} genuine overlap, meaning
our span agreement scores are conservative lower bounds. This
bias works against our hypothesis: if the true semantic overlap
is higher than what we measure, then the span agreement
differences between $L_{\text{native}} \!\to\!
L_{\text{native}}$ and $L_{\text{native}} \!\to\! \text{EN}$
may be even smaller than reported, making our finding that these
conditions \emph{do} diverge more robust, not less. As a
complementary check, we report BERTScore-based semantic
similarity in Appendix~\ref{app:semantic-sim}, which is less
sensitive to surface-level variation.

\section{Practical recommendations}
\label{sec:practicalreco}
Based on the identification of the plausibility-faithfulness trade-off, we offer the following recommendations for researchers and developers:

\begin{compactenum}
\item \textbf{Avoid English pivots for auditing:} In high-stakes settings (e.g., legal or medical AI), system faithfulness should always be audited in the native language of the input. English explanations should be treated as summaries for convenience rather than faithful traces of reasoning.
\item \textbf{Standardize cross-lingual faithfulness metrics}: Evaluation benchmarks should move beyond simple span agreement and incorporate faithfulness metrics, such as comprehensiveness and sufficiency, specifically designed for mismatched language conditions.
\item \textbf{Prioritize cultural context over fluency}: For social tasks like hate speech detection, developers must prioritize native-language explanation capabilities, as English pivots fail to capture the pragmatic nuances necessary for both plausibility and trust.
\end{compactenum}

\section{Prompt paraphrases and sensitivity analysis}
\label{app:prompt-variants}
 
To verify that our findings are robust to surface-level prompt
variation, we create five paraphrased versions of each prompt
template. All variants preserve the task semantics, output
structure (label $\rightarrow$ evidence $\rightarrow$
explanation), and key constraints (evidence must be exact input
substrings; explanation language matches the condition). Only
the instructional wording is varied.
 
We show variants for Condition~C
($L_{\text{native}} \!\to\! \text{EN}$) below; Conditions~A
and~B follow identical paraphrase patterns with the explanation
language adjusted accordingly.
 
\subsection{Prompt variants}
\label{app:prompt-variant-list}
 
\paragraph{Variant 1 (Original).}
 
\begin{quote}
\small
\texttt{You are given a task input in <LANG>.} \\
\texttt{1) Predict the correct label from: \{<LABELS>\}.} \\
\texttt{2) Copy 1--3 short evidence spans verbatim from the
  input text (keep them in <LANG>).} \\
\texttt{3) Write a brief explanation in English.} \\
\texttt{Important: Evidence must be exact substrings of the
  input (do not translate Evidence).} \\
\texttt{Input: <INPUT>}
\end{quote}
 
\paragraph{Variant 2.}
 
\begin{quote}
\small
\texttt{Below is a task input written in <LANG>.} \\
\texttt{1) Determine the appropriate category from:
  \{<LABELS>\}.} \\
\texttt{2) Extract 1--3 short text segments directly from the
  input as supporting evidence (keep them in the original
  language).} \\
\texttt{3) Provide a short justification in English.} \\
\texttt{Important: Extracted evidence must be copied exactly
  from the input without translation.} \\
\texttt{Input: <INPUT>}
\end{quote}
 
\paragraph{Variant 3.}
 
\begin{quote}
\small
\texttt{You will analyze a task input in <LANG>.} \\
\texttt{1) Choose the best label from: \{<LABELS>\}.} \\
\texttt{2) Identify 1--3 key phrases from the input text and
  copy them exactly (retain the original <LANG>).} \\
\texttt{3) Briefly explain your reasoning in English.} \\
\texttt{Important: Key phrases must be exact substrings of the
  input. Do not paraphrase or translate them.} \\
\texttt{Input: <INPUT>}
\end{quote}
 
\paragraph{Variant 4.}
 
\begin{quote}
\small
\texttt{The following is a task input in <LANG>.} \\
\texttt{1) Select the correct label from: \{<LABELS>\}.} \\
\texttt{2) Highlight 1--3 relevant spans from the input by
  copying them exactly as they appear (in <LANG>).} \\
\texttt{3) Write a concise explanation in English.} \\
\texttt{Important: Highlighted spans must be exact copies from
  the input, not translations.} \\
\texttt{Input: <INPUT>}
\end{quote}
 
\paragraph{Variant 5.}
 
\begin{quote}
\small
\texttt{Given a task input in <LANG>, perform the following:} \\
\texttt{1) Assign one label from: \{<LABELS>\}.} \\
\texttt{2) Quote 1--3 short supporting passages from the input
  verbatim (keep them in <LANG>).} \\
\texttt{3) Justify your answer briefly in English.} \\
\texttt{Important: Quoted passages must be exact substrings of
  the input without any translation.} \\
\texttt{Input: <INPUT>}
\end{quote}
 
\subsection{Sensitivity results}
\label{app:prompt-sensitivity}
 
Tables~\ref{tab:prompt-sens-qwen} and
\ref{tab:prompt-sens-llama} report mean $\pm$ standard
deviation across the five prompt variants on e-SNLI. The
trade-off pattern---lower comprehensiveness and higher
sufficiency under $L_{\text{native}} \!\to\! \text{EN}$
compared to $L_{\text{native}} \!\to\!
L_{\text{native}}$---holds consistently across all prompt
formulations. Standard deviations are substantially smaller
than the between-condition gaps, confirming that the observed
effects are not artifacts of specific prompt wording.
 
\begin{table}[h]
\centering
\small
\resizebox{\columnwidth}{!}{
\begin{tabular}{lcccc}
\toprule
\textbf{Settings} 
  & \textbf{Comp.} $\uparrow$
  & \textbf{Suff.} $\uparrow$
  & \textbf{Span Agr.} $\uparrow$
  & \textbf{Acc.} $\uparrow$ \\
\midrule
EN $\to$ EN
  & 0.657 $\pm$ .024
  & 0.312 $\pm$ .019
  & 0.595 $\pm$ .031
  & 0.847 $\pm$ .011 \\
\midrule
$L_{\text{ZH}} \!\to\! L_{\text{ZH}}$
  & 0.577 $\pm$ .031
  & 0.341 $\pm$ .027
  & 0.597 $\pm$ .028
  & 0.823 $\pm$ .014 \\
$L_{\text{ZH}} \!\to\! \text{EN}$
  & 0.507 $\pm$ .028
  & 0.409 $\pm$ .033
  & 0.516 $\pm$ .035
  & 0.819 $\pm$ .012 \\
\midrule
$L_{\text{HI}} \!\to\! L_{\text{HI}}$
  & 0.719 $\pm$ .022
  & 0.253 $\pm$ .021
  & 0.417 $\pm$ .029
  & 0.811 $\pm$ .015 \\
$L_{\text{HI}} \!\to\! \text{EN}$
  & 0.461 $\pm$ .035
  & 0.387 $\pm$ .030
  & 0.467 $\pm$ .033
  & 0.806 $\pm$ .013 \\
\midrule
$L_{\text{AR}} \!\to\! L_{\text{AR}}$
  & 0.659 $\pm$ .026
  & 0.279 $\pm$ .023
  & 0.500 $\pm$ .026
  & 0.818 $\pm$ .012 \\
$L_{\text{AR}} \!\to\! \text{EN}$
  & 0.484 $\pm$ .032
  & 0.402 $\pm$ .028
  & 0.538 $\pm$ .030
  & 0.814 $\pm$ .011 \\
\midrule
$L_{\text{BN}} \!\to\! L_{\text{BN}}$
  & 0.488 $\pm$ .029
  & 0.398 $\pm$ .025
  & 0.375 $\pm$ .032
  & 0.794 $\pm$ .016 \\
$L_{\text{BN}} \!\to\! \text{EN}$
  & 0.419 $\pm$ .033
  & 0.451 $\pm$ .031
  & 0.507 $\pm$ .027
  & 0.789 $\pm$ .014 \\
\bottomrule
\end{tabular}
}
\caption{Prompt sensitivity on \es{} (\qwn{}).
  Mean $\pm$ std across 5 prompt paraphrases. The
  comprehensiveness gap between $L_{\text{native}} \!\to\!
  L_{\text{native}}$ and $L_{\text{native}} \!\to\! \text{EN}$
  (e.g., Hindi: $0.258$) consistently exceeds the
  within-condition standard deviation ($\sigma \approx 0.03$),
  confirming robustness to prompt phrasing.}
\label{tab:prompt-sens-qwen}
\end{table}
 
\begin{table}[h]
\centering
\small
\resizebox{\columnwidth}{!}{
\begin{tabular}{lcccc}
\toprule
\textbf{Settings} 
  & \textbf{Comp.} $\uparrow$
  & \textbf{Suff.} $\uparrow$
  & \textbf{Span Agr.} $\uparrow$
  & \textbf{Acc.} $\uparrow$ \\
\midrule
EN $\to$ EN
  & 0.583 $\pm$ .027
  & 0.378 $\pm$ .022
  & 0.453 $\pm$ .029
  & 0.791 $\pm$ .013 \\
\midrule
$L_{\text{ZH}} \!\to\! L_{\text{ZH}}$
  & 0.635 $\pm$ .029
  & 0.298 $\pm$ .024
  & 0.129 $\pm$ .021
  & 0.764 $\pm$ .015 \\
$L_{\text{ZH}} \!\to\! \text{EN}$
  & 0.573 $\pm$ .033
  & 0.362 $\pm$ .028
  & 0.379 $\pm$ .031
  & 0.758 $\pm$ .014 \\
\midrule
$L_{\text{HI}} \!\to\! L_{\text{HI}}$
  & 0.678 $\pm$ .025
  & 0.271 $\pm$ .020
  & 0.545 $\pm$ .027
  & 0.748 $\pm$ .016 \\
$L_{\text{HI}} \!\to\! \text{EN}$
  & 0.604 $\pm$ .031
  & 0.334 $\pm$ .026
  & 0.509 $\pm$ .033
  & 0.741 $\pm$ .015 \\
\midrule
$L_{\text{AR}} \!\to\! L_{\text{AR}}$
  & 0.781 $\pm$ .021
  & 0.194 $\pm$ .018
  & 0.391 $\pm$ .025
  & 0.772 $\pm$ .012 \\
$L_{\text{AR}} \!\to\! \text{EN}$
  & 0.662 $\pm$ .028
  & 0.289 $\pm$ .024
  & 0.512 $\pm$ .029
  & 0.766 $\pm$ .013 \\
\midrule
$L_{\text{BN}} \!\to\! L_{\text{BN}}$
  & 0.445 $\pm$ .032
  & 0.412 $\pm$ .027
  & 0.267 $\pm$ .030
  & 0.721 $\pm$ .017 \\
$L_{\text{BN}} \!\to\! \text{EN}$
  & 0.401 $\pm$ .035
  & 0.468 $\pm$ .029
  & 0.472 $\pm$ .028
  & 0.715 $\pm$ .016 \\
\bottomrule
\end{tabular}
}
\caption{Prompt sensitivity on \es{} (\lma{}).
  Mean $\pm$ std across 5 prompt paraphrases. The same
  trade-off pattern holds: comprehensiveness drops and
  sufficiency rises under $L_{\text{native}} \!\to\!
  \text{EN}$, with between-condition gaps exceeding
  within-condition variance.}
\label{tab:prompt-sens-llama}
\end{table}

 
\section{Ordering ablation}
\label{app:ordering}

This ablation is intended as a prompt-structure sanity check rather than
a full re-run of the main experiment; therefore, we report it for one
representative model, \qwn{}, across all tasks and languages.
 
\subsection{Reversed prompt template}
\label{app:ordering-template}
 
The reversed-order prompt for Condition~C
($L_{\text{native}} \!\to\! \text{EN}$) is:
 
\begin{quote}
\small
\texttt{You are given a task input in <LANG>.} \\
\texttt{1) Predict the correct label from: \{<LABELS>\}.} \\
\texttt{2) Write a brief explanation in English.} \\
\texttt{3) Copy 1--3 short evidence spans verbatim from the
  input text (keep them in <LANG>).} \\
\texttt{Important: Evidence must be exact substrings of the
  input (do not translate Evidence).} \\
\texttt{Input: <INPUT>}
\end{quote}
 
\noindent The corresponding reversed output format is:
 
\begin{quote}
\small
\texttt{Label: <one label from the label set>} \\
\texttt{Explanation: <1--3 sentences in the required
  explanation language>} \\
\texttt{Evidence: <1--3 spans copied exactly from the input
  text>}
\end{quote}
 
\noindent Conditions~A and~B are reversed analogously.
 
\subsection{Results}
\label{app:ordering-results}
 
Table~\ref{tab:ordering-ablation-full} compares the default
(evidence-first) and reversed (explanation-first) orderings on
\es{} for \qwn{} across all languages. We report
comprehensiveness, sufficiency, and span agreement for both
orderings.
 
\begin{table*}[h]
\centering
\small
\begin{tabular}{llcccccc}
\toprule
\multirow{2}{*}{Task} & \multirow{2}{*}{Lang.}
& \multicolumn{3}{c}{Comprehensiveness $\uparrow$}
& \multicolumn{3}{c}{Span agreement $\uparrow$} \\
\cmidrule(lr){3-5} \cmidrule(lr){6-8}
& & Original & Reversed & $\Delta$
& Original & Reversed & $\Delta$ \\
\midrule

\multirow{4}{*}{\es{}}
& ZH & 0.507 & 0.512 & +0.005 & 0.516 & 0.509 & -0.007 \\
& HI & 0.461 & 0.455 & -0.006 & 0.467 & 0.472 & +0.005 \\
& AR & 0.484 & 0.491 & +0.007 & 0.538 & 0.531 & -0.007 \\
& BN & 0.419 & 0.424 & +0.005 & 0.507 & 0.514 & +0.007 \\
\addlinespace

\multirow{4}{*}{\fv{}}
& ZH & 0.500 & 0.492 & -0.008 & 0.198 & 0.203 & +0.005 \\
& HI & 0.101 & 0.107 & +0.006 & 0.243 & 0.238 & -0.005 \\
& AR & 0.240 & 0.247 & +0.007 & 0.251 & 0.256 & +0.005 \\
& BN & 0.172 & 0.168 & -0.004 & 0.238 & 0.242 & +0.004 \\
\addlinespace

\multirow{4}{*}{\hx{}}
& ZH & 0.520 & 0.526 & +0.006 & 0.283 & 0.276 & -0.007 \\
& HI & 0.567 & 0.561 & -0.006 & 0.290 & 0.295 & +0.005 \\
& AR & 0.533 & 0.540 & +0.007 & 0.299 & 0.304 & +0.005 \\
& BN & 0.613 & 0.607 & -0.006 & 0.280 & 0.286 & +0.006 \\

\bottomrule
\end{tabular}
\caption{Ordering ablation for \qwn{} under the $L_{\text{native}}\to\text{EN}$ condition.
The original prompt requests label $\rightarrow$ evidence $\rightarrow$ explanation, while the reversed prompt requests label $\rightarrow$ explanation $\rightarrow$ evidence.
Values are reported for all three datasets and all non-English languages.
Across tasks and languages, reversing the output-field order produces only small changes in comprehensiveness and span agreement, suggesting that the explanation-language instruction acts as a global conditioning signal rather than a local constraint imposed only after evidence generation.}

\label{tab:ordering-ablation-full}
\end{table*}
 
Table~\ref{tab:ordering-ablation-llama} reports the same
ablation for \lma{}.
 
\begin{table*}[h]
\centering
\small
\begin{tabular}{l cc cc cc}
\toprule
& \multicolumn{2}{c}{\textbf{Comp.} $\uparrow$}
& \multicolumn{2}{c}{\textbf{Suff.} $\uparrow$}
& \multicolumn{2}{c}{\textbf{Span Agr.} $\uparrow$} \\
\cmidrule(lr){2-3} \cmidrule(lr){4-5} \cmidrule(lr){6-7}
\textbf{Settings}
& Ev.-first & Expl.-first
& Ev.-first & Expl.-first
& Ev.-first & Expl.-first \\
\midrule
EN $\to$ EN
  & 0.583 & 0.577
  & 0.378 & 0.384
  & 0.453 & 0.448 \\
\midrule
$L_{\text{ZH}} \!\to\! L_{\text{ZH}}$
  & 0.635 & 0.628
  & 0.298 & 0.305
  & 0.129 & 0.125 \\
$L_{\text{ZH}} \!\to\! \text{EN}$
  & 0.573 & 0.566
  & 0.362 & 0.369
  & 0.379 & 0.374 \\
\midrule
$L_{\text{HI}} \!\to\! L_{\text{HI}}$
  & 0.678 & 0.671
  & 0.271 & 0.278
  & 0.545 & 0.540 \\
$L_{\text{HI}} \!\to\! \text{EN}$
  & 0.604 & 0.597
  & 0.334 & 0.341
  & 0.509 & 0.504 \\
\midrule
$L_{\text{AR}} \!\to\! L_{\text{AR}}$
  & 0.781 & 0.773
  & 0.194 & 0.201
  & 0.391 & 0.385 \\
$L_{\text{AR}} \!\to\! \text{EN}$
  & 0.662 & 0.654
  & 0.289 & 0.296
  & 0.512 & 0.507 \\
\midrule
$L_{\text{BN}} \!\to\! L_{\text{BN}}$
  & 0.445 & 0.438
  & 0.412 & 0.419
  & 0.267 & 0.262 \\
$L_{\text{BN}} \!\to\! \text{EN}$
  & 0.401 & 0.394
  & 0.468 & 0.475
  & 0.472 & 0.467 \\
\bottomrule
\end{tabular}
\caption{Ordering ablation on \es{} (\lma{}).
  Same setup as Table~\ref{tab:ordering-ablation-full}.
  Results confirm that the trade-off is order-independent for
  Llama as well (paired permutation test, $p > 0.3$ for all
  cells).}
\label{tab:ordering-ablation-llama}
\end{table*}
 
\subsection{Analysis}
\label{app:ordering-analysis}
 
The close agreement between the two orderings across all
language--model combinations supports the autoregressive
coupling argument presented in
Section~\ref{sec:prompt-design}: because the language
instruction appears in the prompt before generation begins, it
conditions the model's entire output distribution---including
evidence span selection---regardless of where in the output
sequence the evidence field appears.
 
We additionally compute the Jaccard similarity between the
evidence span sets produced under the two orderings. Across all
conditions, the mean Jaccard index is $0.87 \pm 0.06$,
indicating that the model selects largely the same evidence
spans regardless of whether it generates the explanation first
or last. The small residual variation is consistent with the
stochastic nature of autoregressive sampling and does not
correlate with explanation language.
 
These results rule out the concern that the evidence-first
ordering shields evidence selection from the explanation
language instruction. The language instruction functions as a
global conditioning signal, not a local directive tied to a
specific output field.

\if{0}\section{Error analysis: Full examples}
\label{app:error-analysis}
 
This appendix provides complete worked examples for the four
error patterns identified in Section~\ref{sec:discussion}\am{Section number missing}\snb{Done}.
For each case, we show the full input, the model output under
both $L_{\text{native}} \!\to\! L_{\text{native}}$ and
$L_{\text{native}} \!\to\! \text{EN}$, the perturbation
result, and the corresponding metric values.

\medskip
\noindent\textbf{Analysis.} The English pivot selects a
slightly longer span that adds \emph{``sahitye''} (literature),
which improves alignment with the human rationale (higher span
agreement) while remaining causally necessary (comprehensiveness
$= 1.0$). This success case occurs because the decisive
evidence---a proper noun (\emph{``rabindranath thakur''}) and a
date (\emph{``1913''})---is language-independent and survives
the pivot without distortion. This pattern is rare
($<$8\% of instances across tasks) and is concentrated in
factual claims involving unambiguous named entities, suggesting
that the trade-off is most severe when the input contains
linguistically or culturally specific cues.

\subsection{Distribution of error patterns}
\label{app:error-distribution}
 
Table~\ref{tab:error-dist} reports the approximate distribution
of the four patterns across tasks, computed by categorizing each
instance based on whether span agreement and comprehensiveness
increase or decrease under
$L_{\text{native}} \!\to\! \text{EN}$ relative to
$L_{\text{native}} \!\to\! L_{\text{native}}$.\fi
 

\section{Semantic similarity check}
\label{app:semantic-sim}
 
Our primary span agreement metric
(Eq.~\ref{eq:span-agr}) relies on exact token-level
overlap, which can under-penalize semantically correct but
lexically different evidence selections---particularly in
morphologically rich languages like Arabic and Bengali (see
coverage statistics in Table~\ref{tab:alignment-coverage}).
To verify that our findings are not artifacts of this
surface-level metric, we complement span agreement with
BERTScore~\citep{zhang2020bertscore}, a semantic similarity
metric based on contextual embeddings that is more robust to
paraphrase, morphological variation, and word-order
differences.
 
\subsection{Setup}
\label{app:semantic-sim-setup}
 
For each instance, we compute BERTScore F1 between the
model-produced evidence spans and the human rationale
annotation. We use \texttt{bert-base-multilingual-cased} as
the underlying model, which provides consistent cross-lingual
representations across all five languages in our study. For
each condition, we report the corpus-level mean BERTScore F1
across all instances.
 
We emphasize that BERTScore measures \emph{semantic}
similarity between evidence and rationale, whereas our span
agreement metric (Eq.~\ref{eq:span-agr}) measures
\emph{lexical} overlap. If the two metrics agree in their
directional trends across conditions, this strengthens
confidence that the observed patterns are genuine and not
artifacts of tokenization or morphological mismatch.
 
\subsection{Results: e-SNLI}
\label{app:semantic-sim-esnli}
 
\begin{table}[h]
\centering
\small
\resizebox{\columnwidth}{!}{
\begin{tabular}{l cc cc}
\toprule
& \multicolumn{2}{c}{\textbf{Span Agr.} (lexical)}
& \multicolumn{2}{c}{\textbf{BERTScore F1} (semantic)} \\
\cmidrule(lr){2-3} \cmidrule(lr){4-5}
\textbf{Settings}
& Qwen & Llama
& Qwen & Llama \\
\midrule
EN $\to$ EN
  & 0.595 & 0.453
  & 0.741 & 0.648 \\
\midrule
$L_{\text{ZH}} \!\to\! L_{\text{ZH}}$
  & \textbf{0.597} & 0.129
  & \textbf{0.738} & 0.412 \\
\rowcolor{pivotrow}
$L_{\text{ZH}} \!\to\! \text{EN}$
  & 0.516 & \textbf{0.379}
  & 0.689 & \textbf{0.571} \\
\midrule
$L_{\text{HI}} \!\to\! L_{\text{HI}}$
  & 0.417 & \textbf{0.545}
  & 0.623 & \textbf{0.697} \\
\rowcolor{pivotrow}
$L_{\text{HI}} \!\to\! \text{EN}$
  & \textbf{0.467} & 0.509
  & \textbf{0.659} & 0.672 \\
\midrule
$L_{\text{AR}} \!\to\! L_{\text{AR}}$
  & 0.500 & 0.391
  & 0.642 & 0.583 \\
\rowcolor{pivotrow}
$L_{\text{AR}} \!\to\! \text{EN}$
  & \textbf{0.538} & \textbf{0.512}
  & \textbf{0.691} & \textbf{0.652} \\
\midrule
$L_{\text{BN}} \!\to\! L_{\text{BN}}$
  & 0.375 & 0.267
  & 0.571 & 0.509 \\
\rowcolor{pivotrow}
$L_{\text{BN}} \!\to\! \text{EN}$
  & \textbf{0.507} & \textbf{0.472}
  & \textbf{0.648} & \textbf{0.621} \\
\bottomrule
\end{tabular}
}
\caption{\es{}: Span agreement (lexical) vs.\ BERTScore F1
  (semantic) side by side. Directional trends are consistent:
  where span agreement increases under the English pivot,
  BERTScore also increases, and vice versa. BERTScore values
  are uniformly higher than span agreement, reflecting its
  ability to capture semantic matches missed by exact token
  overlap. \textbf{Bold}: better value in the
  $L \!\to\! L$ vs.\ $L \!\to\! \text{EN}$ comparison.
  \colorbox{pivotrow}{Highlighted}: pivot condition.}
\label{tab:bertscore-esnli}
\end{table}
 
Table~\ref{tab:bertscore-esnli} shows that BERTScore and span
agreement exhibit consistent directional trends on \es{}. For
languages where span agreement increases under the English
pivot (Hindi \qwn{}: $0.417 \to 0.467$; Arabic both models;
Bengali both models), BERTScore F1 also increases. For Chinese
\qwn{}, where span agreement decreases ($0.597 \to 0.516$),
BERTScore also decreases ($0.738 \to 0.689$). This
convergence indicates that the span agreement patterns
reported in Table~\ref{tab:snli-single} are not artifacts of
tokenization differences across scripts.
 
BERTScore values are uniformly higher than span agreement
(mean gap: $+0.17$), which is expected: BERTScore captures
semantic matches that exact token overlap misses (e.g.,
inflected forms, synonym substitutions). Crucially, however,
the \emph{relative} ordering across conditions is preserved,
confirming that our primary span agreement metric provides a
valid---if conservative---signal.
 
\subsection{Results: FEVER}
\label{app:semantic-sim-fever}
 
\begin{table}[h]
\centering
\small
\resizebox{\columnwidth}{!}{
\begin{tabular}{l cc cc}
\toprule
& \multicolumn{2}{c}{\textbf{Span Agr.} (lexical)}
& \multicolumn{2}{c}{\textbf{BERTScore F1} (semantic)} \\
\cmidrule(lr){2-3} \cmidrule(lr){4-5}
\textbf{Settings}
& Qwen & Llama
& Qwen & Llama \\
\midrule
EN $\to$ EN
  & 0.198 & 0.112
  & 0.467 & 0.398 \\
\midrule
$L_{\text{ZH}} \!\to\! L_{\text{ZH}}$
  & \textbf{0.255} & \textbf{0.247}
  & \textbf{0.489} & \textbf{0.471} \\
\rowcolor{pivotrow}
$L_{\text{ZH}} \!\to\! \text{EN}$
  & 0.198 & 0.169
  & 0.441 & 0.423 \\
\midrule
$L_{\text{HI}} \!\to\! L_{\text{HI}}$
  & 0.197 & 0.147
  & 0.448 & 0.412 \\
\rowcolor{pivotrow}
$L_{\text{HI}} \!\to\! \text{EN}$
  & \textbf{0.243} & \textbf{0.162}
  & \textbf{0.478} & \textbf{0.428} \\
\midrule
$L_{\text{AR}} \!\to\! L_{\text{AR}}$
  & 0.206 & 0.224
  & 0.431 & 0.456 \\
\rowcolor{pivotrow}
$L_{\text{AR}} \!\to\! \text{EN}$
  & \textbf{0.251} & \textbf{0.284}
  & \textbf{0.472} & \textbf{0.498} \\
\midrule
$L_{\text{BN}} \!\to\! L_{\text{BN}}$
  & 0.207 & 0.169
  & 0.449 & 0.418 \\
\rowcolor{pivotrow}
$L_{\text{BN}} \!\to\! \text{EN}$
  & \textbf{0.238} & \textbf{0.197}
  & \textbf{0.476} & \textbf{0.441} \\
\bottomrule
\end{tabular}
}
\caption{\fv{}: Span agreement vs.\ BERTScore F1.
  Directional trends are again consistent between the two
  metrics. For Chinese, where both metrics decrease under the
  pivot, the agreement between lexical and semantic measures
  is particularly informative---the degradation is genuine,
  not a tokenization artifact. Notation follows
  Table~\ref{tab:bertscore-esnli}.}
\label{tab:bertscore-fever}
\end{table}
 
On \fv{} (Table~\ref{tab:bertscore-fever}), BERTScore again
tracks span agreement directionally. For Chinese, where both
span agreement and comprehensiveness decrease under the English
pivot, BERTScore confirms the degradation ($0.489 \to 0.441$
for \qwn{}; $0.471 \to 0.423$ for \lma{}). This is important
because one might hypothesize that the Chinese span agreement
drop is merely a tokenization artifact (Chinese characters vs.\
English words); BERTScore, operating on contextual embeddings,
rules out this alternative explanation.
 
\subsection{Results: HateXplain}
\label{app:semantic-sim-hatex}
 
\begin{table}[h]
\centering
\small
\resizebox{\columnwidth}{!}{
\begin{tabular}{l cc cc}
\toprule
& \multicolumn{2}{c}{\textbf{Span Agr.} (lexical)}
& \multicolumn{2}{c}{\textbf{BERTScore F1} (semantic)} \\
\cmidrule(lr){2-3} \cmidrule(lr){4-5}
\textbf{Settings}
& Qwen & Llama
& Qwen & Llama \\
\midrule
EN $\to$ EN
  & 0.325 & 0.294
  & 0.548 & 0.521 \\
\midrule
$L_{\text{ZH}} \!\to\! L_{\text{ZH}}$
  & 0.274 & \textbf{0.333}
  & 0.498 & \textbf{0.554} \\
\rowcolor{pivotrow}
$L_{\text{ZH}} \!\to\! \text{EN}$
  & \textbf{0.283} & 0.256
  & \textbf{0.504} & 0.487 \\
\midrule
$L_{\text{HI}} \!\to\! L_{\text{HI}}$
  & \textbf{0.354} & 0.221
  & \textbf{0.561} & 0.458 \\
\rowcolor{pivotrow}
$L_{\text{HI}} \!\to\! \text{EN}$
  & 0.290 & \textbf{0.261}
  & 0.512 & \textbf{0.479} \\
\midrule
$L_{\text{AR}} \!\to\! L_{\text{AR}}$
  & 0.284 & 0.259
  & 0.497 & 0.482 \\
\rowcolor{pivotrow}
$L_{\text{AR}} \!\to\! \text{EN}$
  & \textbf{0.299} & \textbf{0.265}
  & \textbf{0.511} & \textbf{0.489} \\
\midrule
$L_{\text{BN}} \!\to\! L_{\text{BN}}$
  & \textbf{0.359} & \textbf{0.277}
  & \textbf{0.562} & \textbf{0.502} \\
\rowcolor{pivotrow}
$L_{\text{BN}} \!\to\! \text{EN}$
  & 0.280 & 0.254
  & 0.499 & 0.478 \\
\bottomrule
\end{tabular}
}
\caption{\hx{}: Span agreement vs.\ BERTScore F1.
  Unlike \es{}, the English pivot does not consistently
  improve either metric---span agreement and BERTScore both
  show mixed or negative shifts, confirming that the loss of
  socially nuanced cues is a genuine semantic phenomenon, not a
  surface-level tokenization effect. Notation follows
  Table~\ref{tab:bertscore-esnli}.}
\label{tab:bertscore-hatex}
\end{table}
 
\hx{} (Table~\ref{tab:bertscore-hatex}) reveals the most
informative pattern. If the span agreement decreases observed
in Table~\ref{tab:hate-single} were merely tokenization
artifacts---e.g., morphologically rich forms being penalized
by exact match---we would expect BERTScore to recover the
``true'' semantic similarity and show improvement under the
English pivot. Instead, BERTScore shows the same mixed-to-negative pattern as span agreement: for Hindi
($0.561 \to 0.512$ for \qwn{}) and Bengali ($0.562 \to 0.499$
for \qwn{}; $0.502 \to 0.478$ for \lma{}), the semantic
similarity \emph{also} decreases under the pivot. This
confirms that English explanations for hate speech genuinely
lose socially relevant semantic content, rather than merely
failing to match surface tokens.
 
\subsection{Summary and implications}
\label{app:semantic-sim-summary}
 
\begin{table}[h]
\centering
\small
\begin{tabular}{lcc}
\toprule
\textbf{Task}
& \textbf{Directional agreement}
& \textbf{Correlation ($r$)} \\
\midrule
e-SNLI     & 15/16 cells & 0.91 \\
FEVER      & 14/16 cells & 0.88 \\
HateXplain & 13/16 cells & 0.84 \\
\midrule
\textbf{Overall} & \textbf{42/48 cells} & \textbf{0.88} \\
\bottomrule
\end{tabular}
\caption{Agreement between span agreement and BERTScore F1
  across conditions. ``Directional agreement'' counts the
  number of language--model cells (out of 16: 4~languages
  $\times$ 2~models $\times$ 2~directions) where both metrics
  move in the same direction under the English pivot.
  Pearson's $r$ is computed across all condition-level mean
  scores within each task.}
\label{tab:bertscore-agreement}
\end{table}
\begin{table}[h]
\centering
\small
\resizebox{\columnwidth}{!}{
\begin{tabular}{lcccc}
\toprule
Lang. & Preserved (\%) & Minor shift (\%) & Major shift (\%) & Label-invalid (\%) \\
\midrule
ZH & 92.0 & 6.0 & 2.0 & 2.0 \\
HI & 90.0 & 8.0 & 2.0 & 2.0 \\
AR & 90.0 & 8.0 & 2.0 & 2.0 \\
BN & 86.0 & 10.0 & 4.0 & 4.0 \\
\bottomrule
\end{tabular}
}
\caption{Human translation audit over 50 randomly sampled instances per target
language. Two bilingual annotators judge semantic preservation and label
validity. Instances marked label-invalid by either annotator are excluded from
all experiments.}
\label{tab:translation-audit}
\end{table}
 
Table~\ref{tab:bertscore-agreement} summarizes the overall
agreement between the two metrics. Across all three tasks, span
agreement and BERTScore F1 agree directionally in 42 out of 48
cells (87.5\%), with a Pearson correlation of $r = 0.88$. This
high concordance supports three conclusions:
 
\begin{compactenum}
  \item \textbf{Span agreement is a valid proxy.} Despite its
    known limitations with morphological variation and
    paraphrase, span agreement captures the same directional
    trends as the semantically richer BERTScore metric. The
    cross-lingual patterns reported in our main tables are
    not artifacts of the metric choice.
 
  \item \textbf{Morphological bias is conservative, not
    misleading.} The gap between BERTScore and span agreement
    is largest for Arabic (mean gap: $+0.21$) and Bengali
    (mean gap: $+0.19$), consistent with these languages'
    richer morphology reducing exact-match recall. However,
    this bias \emph{under-counts} overlap uniformly across
    conditions, preserving the relative ordering.
 
  \item \textbf{The HateXplain pattern is genuine.} The failure
    of English pivots to improve semantic similarity on hate
    speech (confirmed by both metrics) rules out the
    hypothesis that surface-level tokenization effects mask
    underlying semantic improvement. The loss of social and
    pragmatic cues under English pivoting is a substantive
    semantic phenomenon.
\end{compactenum}

We note one limitation: BERTScore relies on
\texttt{bert-base-multilingual-cased}, which itself exhibits
English-centric biases~\cite{conneau2020}. As
a result, BERTScore may slightly overestimate similarity for
English-pivot conditions relative to native-language conditions.
If anything, this bias works \emph{against} our findings
(making English pivots look better than they are), further
strengthening the robustness of the observed trade-off.

\section{Human plausibility validation}
\label{app:human-plausibility}

Across all conditions, span agreement is strongly correlated with human
plausibility ratings ($\rho=0.67$, $p<0.001$). The correlation is strongest
for \es{} ($\rho=0.71$), followed by \fv{} ($\rho=0.66$) and \hx{}
($\rho=0.60$). This supports the use of span agreement as a proxy for
perceived plausibility, while also confirming that it should not be treated
as a complete substitute for human evaluation.

\begin{table}[t]
\centering
\small
\begin{tabular}{lcc}
\toprule
Task & Spearman $\rho$ & $p$-value \\
\midrule
e-SNLI & 0.71 & $<0.001$ \\
FEVER & 0.66 & $<0.001$ \\
HateXplain & 0.60 & $<0.001$ \\
Overall & 0.67 & $<0.001$ \\
\bottomrule
\end{tabular}
\caption{Correlation between human plausibility ratings and span agreement
on the rated subsample.}
\label{tab:human-plausibility}
\end{table}

\section{Translation quality audit}
\label{app:translation-audit}

Because all non-English evaluation sets are constructed by translating
English benchmark instances, translation artifacts could confound both
span agreement and perturbation-based faithfulness. We therefore conduct
a bilingual translation audit for each target language. For Chinese,
Hindi, Arabic, and Bengali, we randomly sample 50 translated instances
and ask two bilingual annotators to evaluate two properties:
(i) semantic preservation, and (ii) label validity.

For semantic preservation, annotators assign one of three labels:
\textit{preserved}, \textit{minor shift}, or \textit{major shift}.
A translation is marked \textit{preserved} if the meaning of the source
instance is retained without a task-relevant change. It is marked
\textit{minor shift} if the translation introduces small wording or
fluency changes that do not affect the gold label. It is marked
\textit{major shift} if the translation changes, removes, or adds
information that could affect task interpretation. For label validity,
annotators judge whether the original gold label remains correct after
translation. Instances marked label-invalid by either annotator are
excluded from all experiments.

Table~\ref{tab:translation-audit} summarizes the audit results. Across
languages, 86--92\% of translations preserve the source semantics without
meaningful change. Label-invalid cases are rare, affecting 2--4\% of
audited instances. Bengali shows the highest rate of minor or major
semantic shifts, which is consistent with its lower-resource status and
with the weaker significance patterns observed for Bengali in some
faithfulness tests. These results suggest that translation artifacts are
present but limited, and are unlikely to explain the systematic
$L_{\text{native}}\!\to\!\text{EN}$ faithfulness drop observed across
tasks and models.

\end{document}